\documentclass[]{interact}

\usepackage{epstopdf}
\usepackage[caption=false]{subfig}

\usepackage{color}
\usepackage{lmodern}
\usepackage{fixltx2e}

\usepackage[numbers,sort&compress]{natbib}
\bibpunct[, ]{[}{]}{,}{n}{,}{,}
\renewcommand\bibfont{\fontsize{10}{12}\selectfont}
\makeatletter
\def\NAT@def@citea{\def\@citea{\NAT@separator}}
\makeatother

\theoremstyle{plain}

\theoremstyle{definition}

\theoremstyle{remark}

\usepackage{bm}

\newcommand{\etal}{\textit{et al.}}

\begin{document}

\articletype{FULL PAPER}%

\title{Dynamic properties and motion reproducibility of a compact pneumatically actuated humanoid upper body for data-driven control}

\author{
\name{Hiroshi Atsuta\textsuperscript{a}\thanks{CONTACT Hiroshi Atsuta. Email: hiroshi.atsuta@otri.osaka-u.ac.jp}, Hisashi Ishihara\textsuperscript{a,b}, and Minoru Asada\textsuperscript{a,c,d}}
\affil{\textsuperscript{a}Symbiotic Intelligent Systems Research Center, Institute for Open and Transdisciplinary Research Initiatives, The University of Osaka, Suita, Osaka, Japan; \textsuperscript{b}College of Arts and Design, Ritsumeikan University, Kyoto, Japan; \textsuperscript{c}International Professional University of Technology in Osaka, Umeda, Kita-ku, Osaka, Japan; \textsuperscript{d}Chubu University Academy of Emerging Sciences, Kasugai, Aichi, Japan}
}

\maketitle

\begin{abstract} 

  Pneumatically-actuated anthropomorphic robots with high degrees of freedom (DOF) offer significant potential for physical human-robot interaction. However, precise control of pneumatic actuators is challenging due to their inherent nonlinearities. This paper presents the development of a compact 13-DOF upper-body humanoid robot.
  To assess the feasibility of an effective controller, we first investigate its key dynamic properties, such as actuation time delays, and confirm that
  its behavior is reproducible across repeated trials. Leveraging this reproducibility, we implement a preliminary data-driven controller for a 4-DOF arm subsystem based on a multilayer perceptron with explicit time delay compensation. The network was trained on random movement data to generate pressure commands for tracking arbitrary trajectories. In comparative evaluations on this subsystem, the data-driven controller achieved lower tracking errors than a traditional PID controller. This result suggests that data-driven approaches are a promising option for controlling complex, high-DOF pneumatic robots.
\end{abstract}

\begin{keywords}
pneumatic robot; data-driven control; time-delay compensation; trajectory tracking; inverse dynamics; human-robot interaction; humanoid; android
\end{keywords}


\section{Introduction}
\label{sec:introduction}

For a human-robot symbiotic society to be established, physical human-robot interaction (HRI) involving close contact is crucial. Pneumatic actuators are effective for safe physical interaction due to their inherent compliance \cite{ikemoto.etal2012}, which stands in contrast to electrically-driven actuators that use high-transmission-ratio gears \cite{ogenyi.etal2021}.
Pneumatic actuators generally fall into two categories: pneumatic artificial muscles (PAMs) and pneumatic cylinders. PAMs are lightweight and flexible, making them suitable for dynamic movements like jumping and walking in legged robots \cite{hosoda.etal2010,vanderborght.etal2008} and for direct interaction with humans \cite{shin.etal2010,niiyama2022}. However, they require sophisticated control systems to handle their complex dynamics, and their relatively short fatigue life limits their durability \cite{klute.hannaford1998,woods.etal2012}. On the other hand, pneumatic cylinders have more rigid structures and simpler dynamics than PAMs, resulting in longer lifespans, lower maintenance, and better controllability. They offer high force output and compliance but require sturdy metal housings and piston rods.

Designing a compact, pneumatically-actuated robot that replicates the human musculoskeletal system is challenging but essential for effective physical HRI. Furthermore, compactness also facilitates emotional HRI \cite{billard.etal2007,dautenhahn.etal2009}, as smaller robots is often less intimidating and helps people feel more safe and comfortable. Ishihara \etal \cite{ishihara.asada2015} addressed this by developing a compact upper-body robot with 22 degrees of freedom (DOF) driven by pneumatic semi-rotary actuators and cylinders, all within a small frame of approximately 30 cm from waist to shoulders, as part of the ``Affetto'' project \cite{ishihara.etal2011}. They achieved this compactness by carefully selecting joint mechanisms that optimized the use of limited internal space while meeting power requirements.
This previous work demonstrated the feasibility of a small, high-DOF upper body by strategically employing pneumatic semi-rotary actuators and cylinders with direct drive, slider-crank, and parallel linkage mechanisms to ensure both sufficient power and compactness.
While multi-DOF robotic systems with pneumatic cylinders are less common than PAM-driven robots, several have been reported, including robotic arms \cite{noritsugu.park1994,hoshino2008,murayama.etal2014,hoffmann.etal2021}, legged robots \cite{binnard1995,wait.goldfarb2014}, and humanoids \cite{minato.etal2007,todorov.etal2010,tassa.etal2013,tanaka.etal2021a}. However, to the best of the authors' knowledge, no other humanoid robot driven by pneumatic cylinders has achieved such compactness \cite{ishihara.asada2015}.

A fundamental HRI task is kinesthetic teaching, where a robot is physically guided through a motion that it must then reproduce. To achieve this, the ability to track an arbitrary recorded trajectory is essential. However, precisely controlling pneumatic actuators remains challenging due to their nonlinearities, including pressure dynamics, air compressibility, and friction \cite{beater2007,jamian.etal2020}. Numerous control approaches have been proposed, ranging from linear methods augmented with schemes like fuzzy logic \cite{parnichkun.ngaecharoenkul2001,hosovsky.etal2012} and gain scheduling \cite{hamiti.etal1996,situm.etal2004} to nonlinear methods employing more accurate models, such as sliding mode control \cite{song.ishida1997,bouri.thomasset2001,tsai.huang2008,smaoui.etal2008} and adaptive backstepping control \cite{rahman.etal2016,ayadi.etal2018,ren.etal2019}. These studies, however, typically focus on single-DOF experimental setups and require extensive system modeling. A further challenge, specific to HRI contexts, is the need for long air transmission lines to situate the robot away from bulky valves and noisy compressors, which introduces significant time delays and mass flow attenuation \cite{yang.etal2011,turkseven.ueda2018,butt.sepehri2019}. As reported in a previous study \cite{ishihara.asada2015}, a simple linear controller could not achieve adequate tracking performance for such a system.

This study explores the potential of a data-driven approach for controlling such a complex robot, motivated by the challenges of identifying a complete dynamic model. This approach requires both system reproducibility and a sufficient amount of data. The robot in the previous study \cite{ishihara.asada2015} had insufficient actuator power and structural stiffness, leading to unreliable motion generation over long periods. Therefore, we improved the mechanical design of the robot to address these shortcomings in collaboration with A-Lab Co., Ltd. (Tokyo, Japan). Consequently, we investigate the robot's dynamic characteristics, including transmission line time delays, minimum pressures to start movements, and motion characteristics under maximum pressure differences, and examine its reproducibility under specific conditions. Furthermore, we confirm the robot's durability for collecting time-series sensor data over extended periods. Finally, after confirming reproducibility across the 13-DOF platform, we implement a preliminary data-driven controller for a representative 4-DOF arm subsystem as a feasibility study of explicit time-delay compensation, and compare its tracking performance against that of a traditional PID controller.

\begin{figure}[t]
  \begin{center}
    \includegraphics[width=14cm,bb=0 0 503 337]{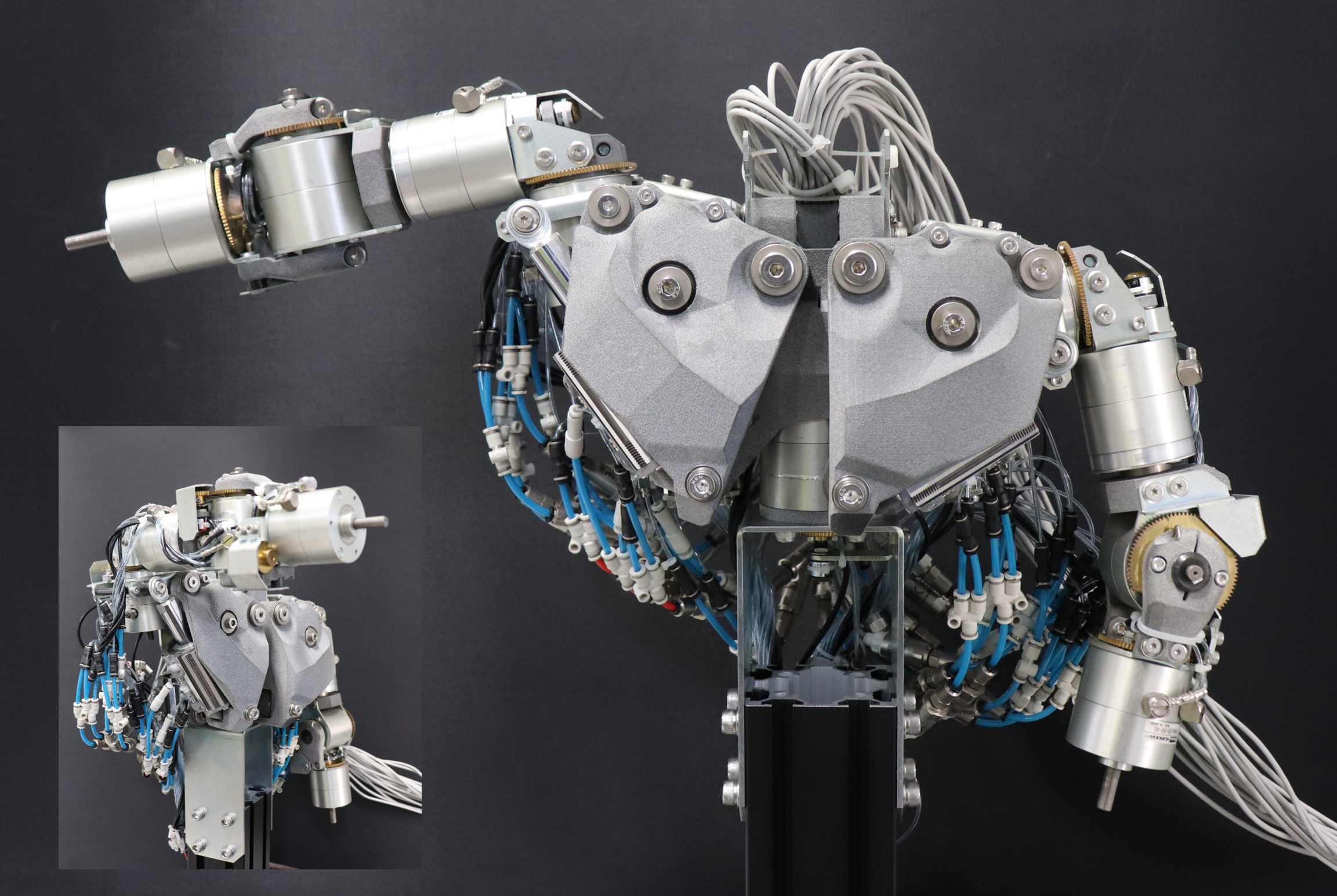}
  \end{center}
  \caption{The 13-DOF pneumatically-actuated upper-body humanoid robot developed in this study. The main photo shows the front view while the inset on the lower left shows the robot's front-right side with a different pose.}
  \label{Affetto}
\end{figure}

This paper is structured as follows: Section~\ref{sec:robot-system} details the design of the compact upper-body structure shown in Figure~\ref{Affetto}. Section~\ref{sec:dynamic-properties} investigates the robot's dynamic properties, focusing on time delays, motion characteristics, and reproducibility. Section~\ref{sec:data-driven-control} describes the data collection process and demonstrates the data-driven controller with explicit time delay compensation. Section~\ref{sec:discussion} discusses the robot's potential for dynamic expression, the limitations of the present study, and future directions for data-driven control. Finally, Section~\ref{sec:conclusion} concludes the paper.


\section{Robot System}
\label{sec:robot-system}

The design policy of this robot follows the ``Affetto'' concept discussed in the preceding work \cite{ishihara.asada2015}. The robot was developed for close physical human-robot interaction with a child-sized upper body, and this led to three main design requirements. First, pneumatic actuation was adopted for physical interaction. Second, the mechanism had to remain compact while preserving a wide motion range for expressive upper-body movements. Third, the noisy pneumatic pump and related equipment were placed away from the robot so as not to disturb interaction, which required long transmission tubes. As discussed later, these long tubes also introduce delay and attenuation in pressure transmission, which directly shape the control challenges addressed in this paper.

\subsection{Mechanical System}
\label{sec:mechanical-system}
\begin{figure}[t]
  \centering
  \includegraphics[width=14cm,bb=0 0 613 475]{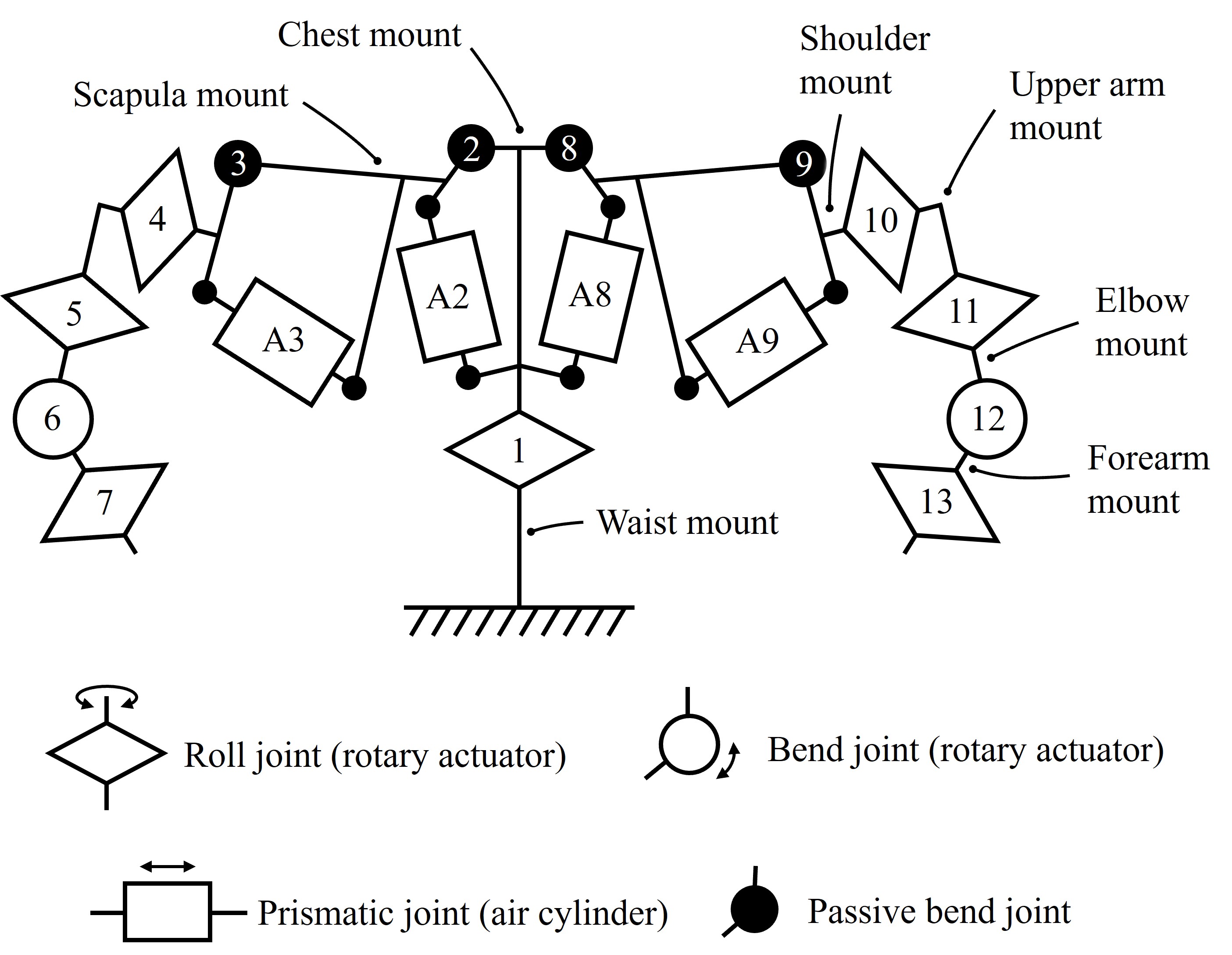}
  \caption{Kinematic diagram of the 13-DOF upper body. Joints 2, 3, 8, and 9 in the chest and scapula are driven by air cylinders A2, A3, A8, and A9, while the others driven by rotary actuators.}
  \label{fig:jointlocation}
\end{figure}

Figure~\ref{fig:jointlocation} illustrates the joint structure of the upper-body robot, which has a total of 13 active joints assembled with seven types of mounts. The rotary actuator for waist rotation (joint 1) is fixed to the waist mount and rotates the chest mount. Two air cylinders for scapula rotation (joints 2 and 8) are fixed to the chest mount to rotate the left and right scapula mounts, while two other air cylinders for shoulder abduction (joints 3 and 9) are fixed to the scapula mounts to rotate the shoulder mounts. These joints must lift or support the arm against gravity within the limited space available in the chest and scapula. For this reason, closed-loop slider-crank mechanisms were adopted so that the force of linear cylinders could be used efficiently while keeping the shoulder structure compact and maintaining the required range of motion. In the arms, the rotary actuators for shoulder flexion (joints 4 and 10) are fixed to the shoulder mounts to rotate the upper arm mounts, and the actuators for shoulder rotation (joints 5 and 11) are fixed to the upper arm mounts to rotate the elbow mounts. The actuators for elbow flexion (joints 6 and 12) are fixed to the elbow mounts to rotate the forearm mounts, which carry the final rotary actuators for forearm rotation (joints 7 and 13). For this robot, the positive direction for each joint is defined as a movement that extends the limbs or opens the body, while the negative direction corresponds to a movement of flexing or folding the limbs.

\begin{figure}
  \centering
  \includegraphics[width=14cm,bb=0 0 941 573]{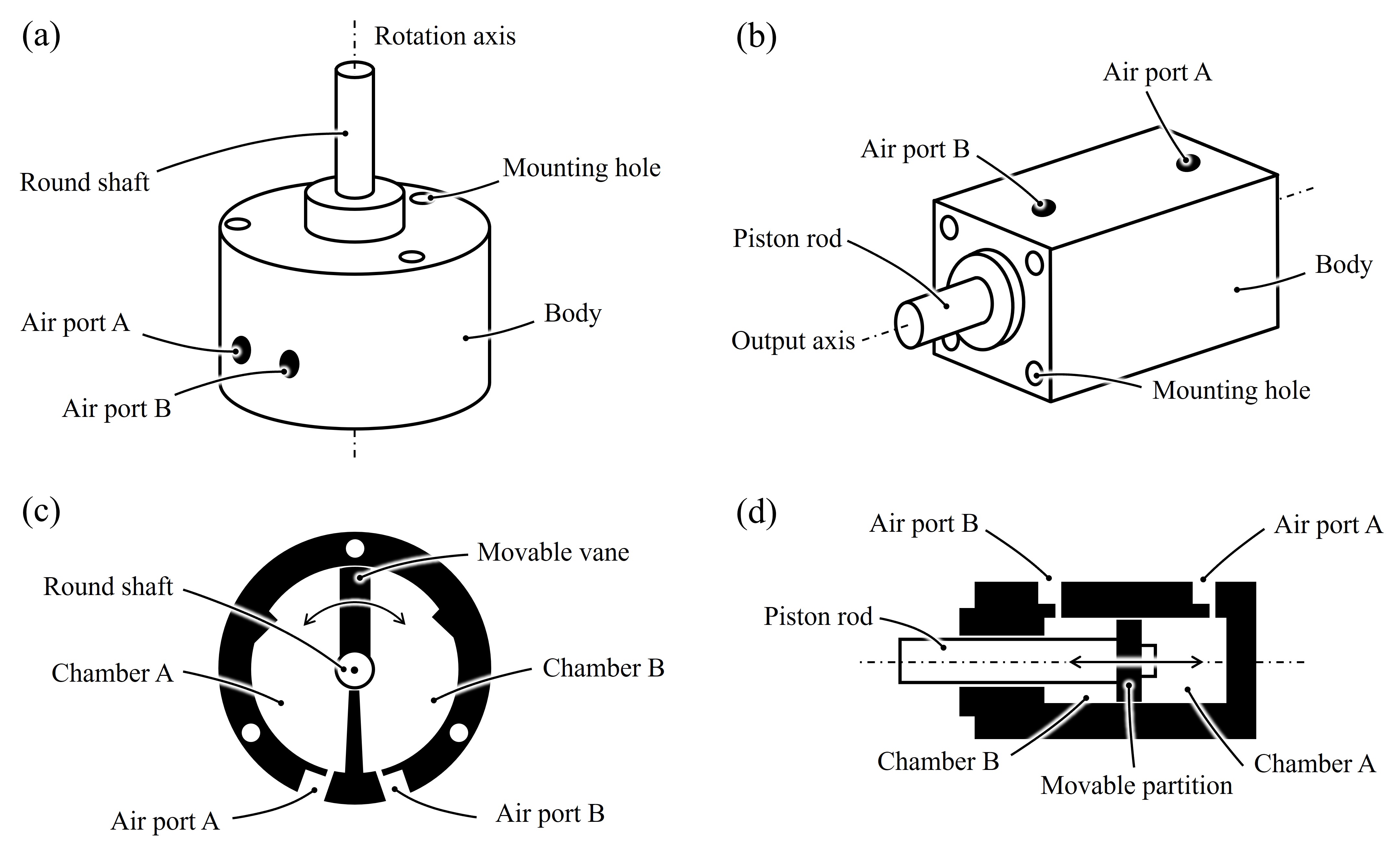}
  \caption{Actuator mechanisms. (a) Assembly drawing of a vane-type rotary actuator. (b) Assembly drawing of an air cylinder. (c) Internal schematic of the rotary actuator. (d) Internal schematic of the air cylinder.}
  \label{fig:actuator}
\end{figure}

Figures \ref{fig:actuator}(a) and \ref{fig:actuator}(b) show the vane-type rotary actuators and the air cylinders used in the robot, respectively. These actuators were customized by A-Lab Co., Ltd. from commercial products, whose specific off-the-shelf type numbers of pneumatic actuators (both the air cylinders and vane-type rotary actuators) are industrial secrets of A-Lab. Both actuator types have rigid bodies with mounting holes for attachment. Each actuator has two air ports (A and B) for the pneumatic tubes. As shown in the schematics in Figures \ref{fig:actuator}(c) and \ref{fig:actuator}(d), each actuator contains two air chambers (A and B). Pressurized air in these chambers pushes a movable vane (rotary actuator) or a movable partition (air cylinder). The resulting torque (Figure \ref{fig:actuator}(c)) or force (Figure \ref{fig:actuator}(d)) is determined by both the pressure difference between these chambers and the pressure-receiving surface area of the vane or partition. Therefore, controlling the pressure in these two chambers is crucial for motion control.

\begin{figure}[t]
  \centering
  \includegraphics[width=14cm,bb=0 0 921 696]{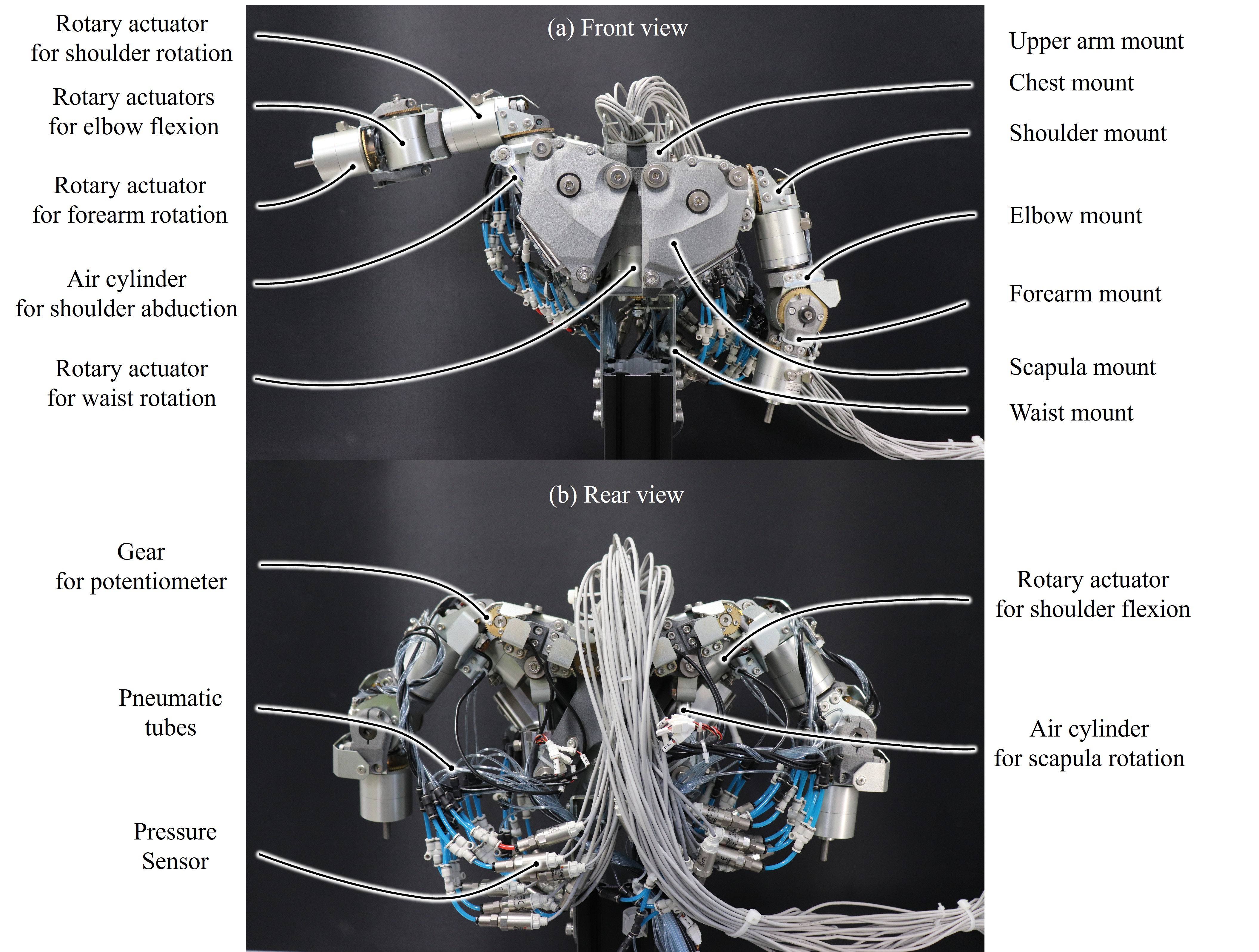}
  \caption{Hardware overview of the developed robot: (a) front view and (b) rear view, with key components labeled.}
  \label{fig:hardware}
\end{figure}

The assembled robot is shown from the front and rear in Figure~\ref{fig:hardware}, where key actuators and structural mounts are indicated. The rear view reveals a complex tubing system where each actuator is connected via two pneumatic tubes, approximately $2.5$~m in length, which allows for remote placement of the control valves. These tubes have a primary internal diameter of $1$~mm, expanding to $2$~mm at the branch section used to connect a pressure sensor (Nihon Pisco Co., Ltd., PSE530) near each actuator's air port. To measure joint angles, potentiometers (NIDEC COMPONENTS CORPORATION, JC10) are installed at each joint. The robot's height from the waist mount is approximately $38$~cm, and its mass is $5.12$~kg.

Table~\ref{tab:specification} summarizes the mechanical specifications of the actuators installed for the 13 active joints. The table lists the corresponding joint number, function, actuator type, rated output at $0.7$~MPaG, and range of motion for each joint. Because the output forces of the air cylinders differ according to the movement direction of the piston rod, the rated outputs for the pulling movement are described in parentheses after those for the pushing movement.

\begin{table}[t]
  \footnotesize
  \centering
  \caption{Mechanical specifications of the actuators for the 13 active joints.}
  \label{tab:specification} 
  \begin{tabular}{c c c c c}
    \hline \hline
    \textbf{Joint No.} & \textbf{Function} & \textbf{Type} & \textbf{Output} & \textbf{Range} \\
    \hline
    1 & Waist rotation & Rotary & 5.5 N$\cdot$m & 90 deg \\ 
    2 & Right scapula rotation & Cylinder &  525 (420) N& 30 mm \\ 
    3 & Right shoulder abduction & Cylinder & 350 (260) N& 15 mm \\
    4 & Right shoulder flexion & Rotary & 3.0 N$\cdot$m & 90 deg \\
    5 & Right shoulder rotation & Rotary & 3.0 N$\cdot$m & 90 deg \\ 
    6 & Right elbow flexion & Rotary & 3.0 N$\cdot$m & 90 deg \\ 
    7 & Right forearm rotation & Rotary & 3.0 N$\cdot$m & 90 deg \\
    8 & Left scapula rotation & Cylinder &   525 (420) N& 30 mm \\ 
    9 & Left shoulder abduction & Cylinder & 350 (260) N& 15 mm \\
    10 & Left shoulder flexion & Rotary & 3.0 N$\cdot$m & 90 deg \\
    11 & Left shoulder rotation & Rotary & 3.0 N$\cdot$m & 90 deg \\ 
    12 & Left elbow flexion & Rotary & 3.0 N$\cdot$m & 90 deg \\ 
    13 & Left forearm rotation & Rotary & 3.0 N$\cdot$m & 90 deg \\
    \hline \hline
  \end{tabular}
\end{table}

\subsection{Drive, Sensory, and Control Systems}
\label{sec:drive-sensory-control}

\begin{figure}[t]
  \centering
  \includegraphics[width=0.8\linewidth]{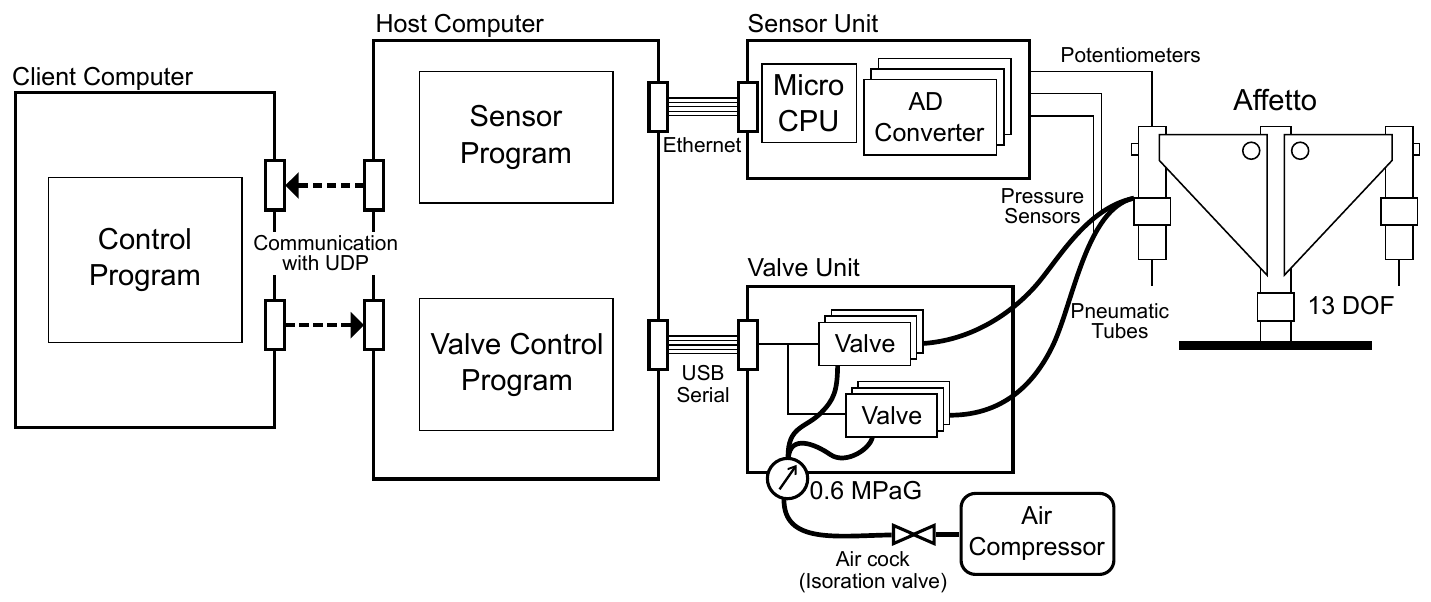}
  \caption{Overview of the experimental control system. A host computer manages valve commands and sensor data acquisition, allowing a client computer to interface with the robot over a local network.}
  \label{fig:system_overview}
\end{figure}

As illustrated in Figure~\ref{fig:system_overview}, the robot is controlled by a host computer that communicates with the hardware via two main channels. First, the valve unit, containing a series of proportional pressure control valves, is connected via USB serial and regulated by an air compressor at $0.6$~MPaG. Second, the sensor unit, containing a micro CPU with analog-to-digital (AD) converters, is linked to the robot's 13 potentiometers and 26 pressure sensors, which sends sensor data to the host computer via Ethernet. Programs for valve control and sensor acquisition run as separate processes on the host computer, while a client computer can interface with the host computer over a local network via UDP to issue commands and log data.

The distal end of each pneumatic tube is connected to a proportional pressure control valve, which discharges compressed air at a regulated pressure from approximately $0.003$ to $0.6$~MPaG with a low hysteresis of $0.25{\%}$ and high precision of $0.75{\%}$. The pressure setpoint is operated by an input voltage from $0$ to $10$~V proportionally. For example, the valve discharges $0$, $0.3$, and $0.6$~MPaG when the input voltage is set to $0$, $5$, and $10$~V, respectively.

The input voltage for each valve is controlled by a one-byte integer command value ($0$--$255$), which is converted to an actual voltage from $0$ to $10$~V by a digital-to-analog converter. Consequently, the control signal for the robot is a $26$-dimensional vector of one-byte elements, since the robot has 13 actuators, each with two pneumatic tube systems. This control signal is updated in the host computer and sent to the robot at $30$~Hz.

The robot has a total of 13 potentiometers and 26 pressure sensors to measure the angles of active joints and the air pressures near the actuator chambers. The reference voltage for these sensors is $5$~V, and their output voltages from $0$ to $5$~V are read by a $16$-bit AD converter (CONTEC CO., LTD., ADI16-4(FIT)GY). These sensory signals are acquired by the host computer at $30$~Hz.

\subsection{Posture Variation}
\label{sec:posture-variation}

\begin{figure}[t]
  \centering
    \includegraphics[width=14cm,bb=0 0 882 468]{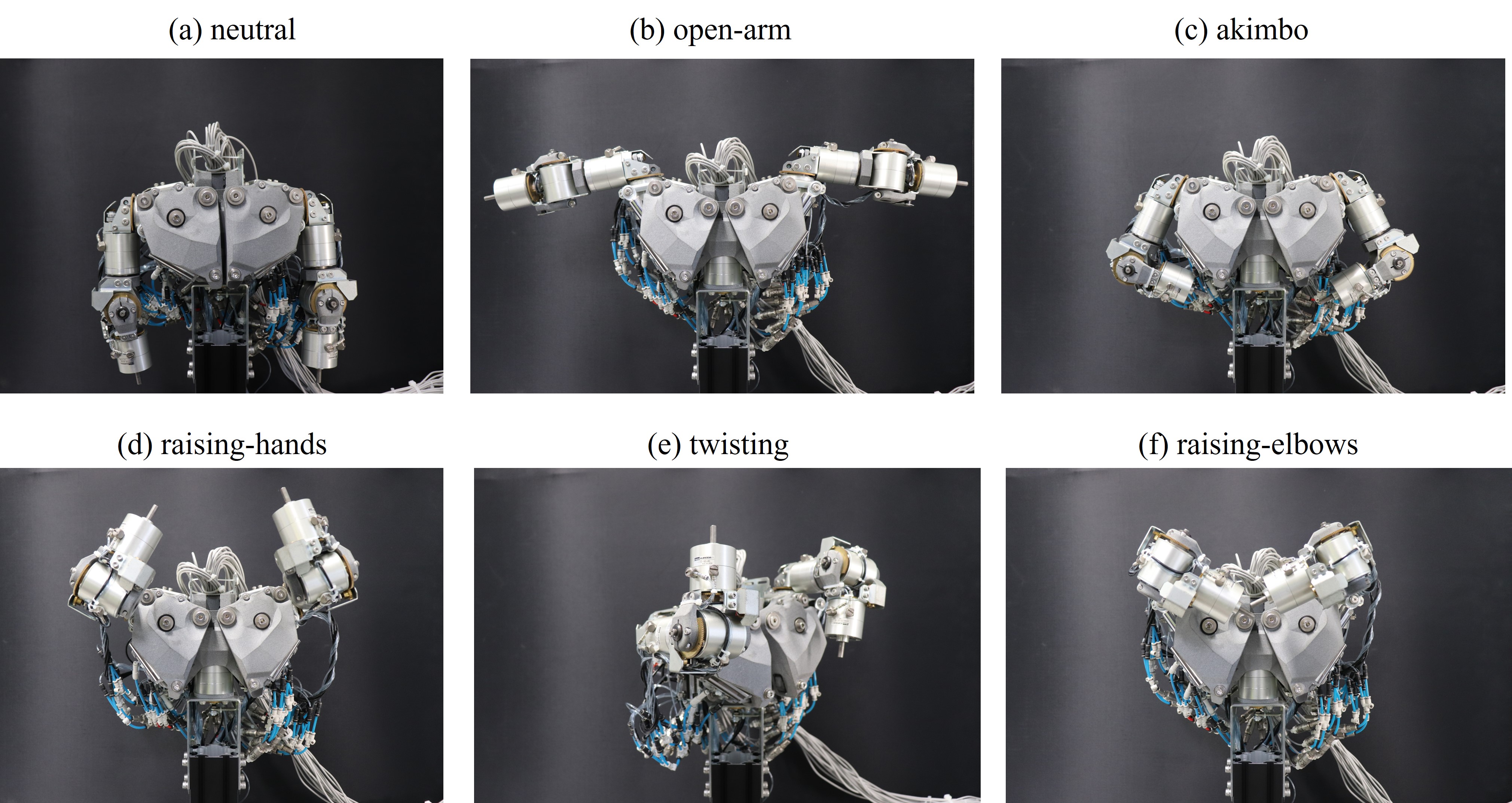}
  \caption{Examples of the robot's expressive postures.}
  \label{fig:poses}
\end{figure} 

Figure~\ref{fig:poses} shows examples of postures that the robot can achieve, including (a) a neutral pose, in which every joint is relaxed; (b) an open-arm pose, in which both arms are spread out; (c) an akimbo pose, in which both hands are set to the waist; (d) a raising-hands pose, in which the hands are raised toward the expected head area; (e) a twisting pose, in which the arms are twisted in opposite directions; and (f) a raising-elbows pose, in which both elbows are raised with folded hands.

\section{Dynamic Properties and Motion Reproducibility}
\label{sec:dynamic-properties}

To design an effective controller for the robot's complex pneumatic system, it is essential to first understand its fundamental dynamic characteristics. This section details a series of experiments conducted to probe these properties. The investigation is divided into two main parts:
\begin{itemize}
\item \textbf{Experiment I} characterizes three key dynamic properties: transmission time delay, the minimum pressure command required for activation, and maximum joint velocity. These tests are designed to highlight the system's nonlinearities and the challenges involved in developing an analytical model.
\item \textbf{Experiment II} assesses the trial-to-trial reproducibility of the robot's movements. This test is crucial for validating the feasibility of a data-driven control approach, which relies on the system behaving deterministically under consistent conditions.
\end{itemize}

\subsection{Experimental Setup}
\label{sec:experimental-setup}

\subsubsection{Tested Joints and Postures}
\begin{figure}[t]
  \centering
  \includegraphics[width=0.7\linewidth]{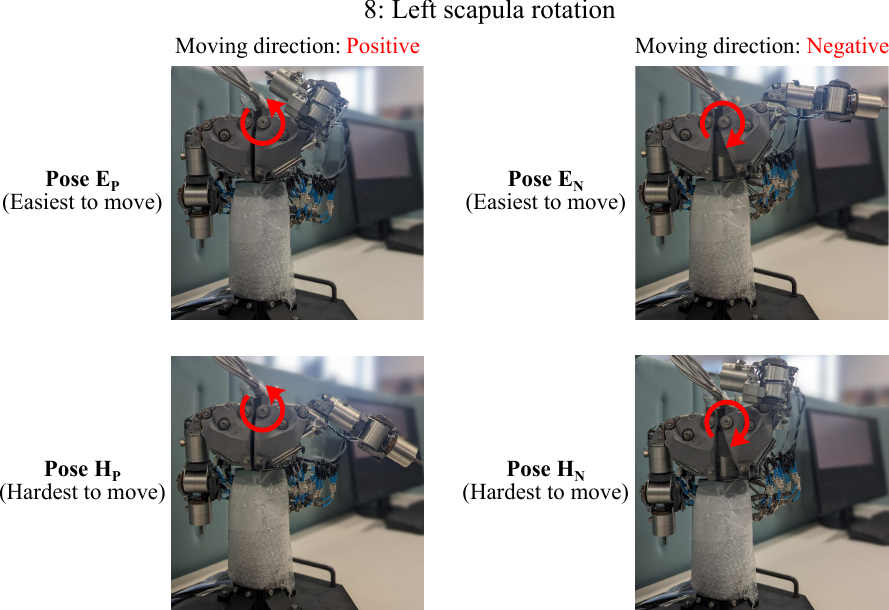}
  \caption{Example of test postures for the left scapula rotation joint (joint 8). The physical posture for the 'easiest' and 'hardest' conditions depends on the movement direction relative to gravity. For positive movement (left), the folded-arm posture is defined as Pose E\textsubscript{P} (easiest) and the extended-arm posture as Pose H\textsubscript{P} (hardest). For negative movement (right), where gravity assists the motion, the extended-arm posture becomes the 'easiest' condition (Pose E\textsubscript{N}), while the folded-arm posture becomes the 'hardest' (Pose H\textsubscript{N}).}
  \label{fig:postures}
\end{figure}

Due to the symmetric design of the robot's arms, experiments were performed on a representative set of 7 joints: the waist (joint 1) and the six joints of the left arm (scapula rotation, shoulder abduction, shoulder flexion, shoulder rotation, elbow flexion, and forearm rotation; joints 8-13). The robot's dynamic response is highly dependent on its posture. Specifically, the gravitational and inertial loads on any single joint under test vary significantly based on the configuration of its distal links (e.g., the shoulder joint experiences a much higher load when the arm is fully extended compared to when it is folded). To systematically investigate the effect of the configuration on the movement, four specific initial postures were defined for each test, accounting for the movement direction (Positive or Negative). These postures were chosen as the extremes of the load range for a given joint:
\begin{itemize}
\item For positive direction movements:
  \begin{itemize}
  \item \textbf{Pose E\textsubscript{P} (Easiest-to-move, Positive)}: The robot is configured to minimize the gravitational and inertial load resisting a given positive movement.
  \item \textbf{Pose H\textsubscript{P} (Hardest-to-move, Positive)}: The robot is configured to maximize the load resisting a given positive movement.
  \end{itemize}
\item For negative direction movements:
  \begin{itemize}
  \item \textbf{Pose E\textsubscript{N} (Easiest-to-move, Negative)}: The robot is configured to minimize the load resisting a given negative movement.
  \item \textbf{Pose H\textsubscript{N} (Hardest-to-move, Negative)}: The robot is configured to maximize the load resisting a given negative movement.
  \end{itemize}
\end{itemize}
Thus, these ``easiest'' and ``hardest'' postures are defined relative to the specific joint being tested. For example, to test the left scapula rotation in the positive direction (lifting the arm against gravity), Pose E\textsubscript{P} involves folding the arm (changing the posture of distal links) to minimize the moment arm, while Pose H\textsubscript{P} involves extending the arm to maximize the load, as shown in Figure~\ref{fig:postures}. Conversely, for negative-direction movements where gravity assists the motion, these physical postures are reversed: Pose E\textsubscript{N} (easiest) is the extended-arm posture, and Pose H\textsubscript{N} (hardest) is the folded-arm posture.

\subsubsection{Notation}
\label{sec:notation}
In this paper, joint angles are expressed as normalized positions in percent of each joint's range of motion, where $0$ and $100$ correspond to the two mechanical limits of the joint, and joint velocities are given in this normalized unit per second. Throughout this section, we use the following notation: $\bm{u}^{\mathrm{A}}(t)$ and $\bm{u}^{\mathrm{B}}(t)$ represent the vectors of commanded values for the A and B actuator chambers at time $t$, respectively. Similarly, $\bm{p}^{\mathrm{A}}(t)$ and $\bm{p}^{\mathrm{B}}(t)$ are the measured pressure values near each chamber, $\bm{q}(t)$ is the vector of measured joint angles, and $\bm{v}(t)$ is the vector of joint velocities (numerical derivative of $\bm{q}(t)$). All vectors are 7-dimensional (e.g., $\bm{u}^{\mathrm{A}}(t)\in \mathbb{R}^{7}$), corresponding to the 7 joints under test. The subscript $i$ indicates the value for the $i$-th joint (e.g., $q_{i}(t)$).

\subsection{Experiment I-A: Measuring Transmission Time Delay}
\label{sec:exp-time-delay}

\subsubsection{Method}

\begin{figure}[t]
  \centering
  \subfloat[Positive direction]{%
    \resizebox*{0.49\linewidth}{!}{%
      \includegraphics{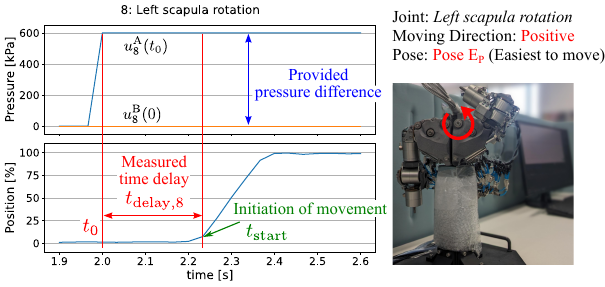}}}
  \label{fig:measure_time_delay_a}
  \subfloat[Negative direction]{%
    \resizebox*{0.49\linewidth}{!}{%
      \includegraphics{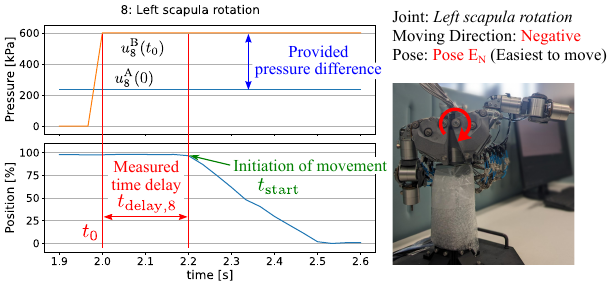}}}
  \label{fig:measure_time_delay_b}
  \caption{Method for measuring time delay ($t_{\mathrm{delay},8}$) for the left scapula rotation joint. The delay is the duration between the command step input at $t_{0}$ and the detected initiation of movement at $t_{\mathrm{start}}$.}
  \label{fig:measure_time_delay}
\end{figure}

A significant time delay is expected between the onset of a valve command and the resulting pressure change in the actuator chamber due to the 2.5 m long pneumatic tubes. This experiment quantifies this delay. For each joint, direction, and posture, the following procedure was repeated 10 times:
\begin{enumerate}
\item To ensure a consistent starting condition, the robot was set to its initial posture (Pose E\textsubscript{P}, H\textsubscript{P}, E\textsubscript{N} or H\textsubscript{N}) by driving the target joint to one of its mechanical limits. This position was held with an initial set of valve commands, $\bm{u}^{\mathrm{A}}(0)$ and $\bm{u}^{\mathrm{B}}(0)$.
\item At time $t_{0}$, a step input was applied to one chamber's valve command (e.g., $u^{\mathrm{A}}_{i}(t_{0})$) to induce movement, while the other command ($u^{\mathrm{B}}_{i}(0)$) was held constant.
\item The joint angle $q_{i}(t)$ and commanded pressures $u^{\mathrm{A}}_{i}(t)$ and $u^{\mathrm{B}}_{i}(t)$ were recorded.
\item The time when movement was first detected, $t_{\mathrm{start}}$, was identified from the joint angle data.
\item The time delay was calculated as $t_{\mathrm{delay},i} = t_{\mathrm{start}} - t_{0}$, as depicted in Figure~\ref{fig:measure_time_delay}.
\end{enumerate}

\subsubsection{Results}

\begin{figure}[t]
  \centering
  \includegraphics[width=.6\linewidth]{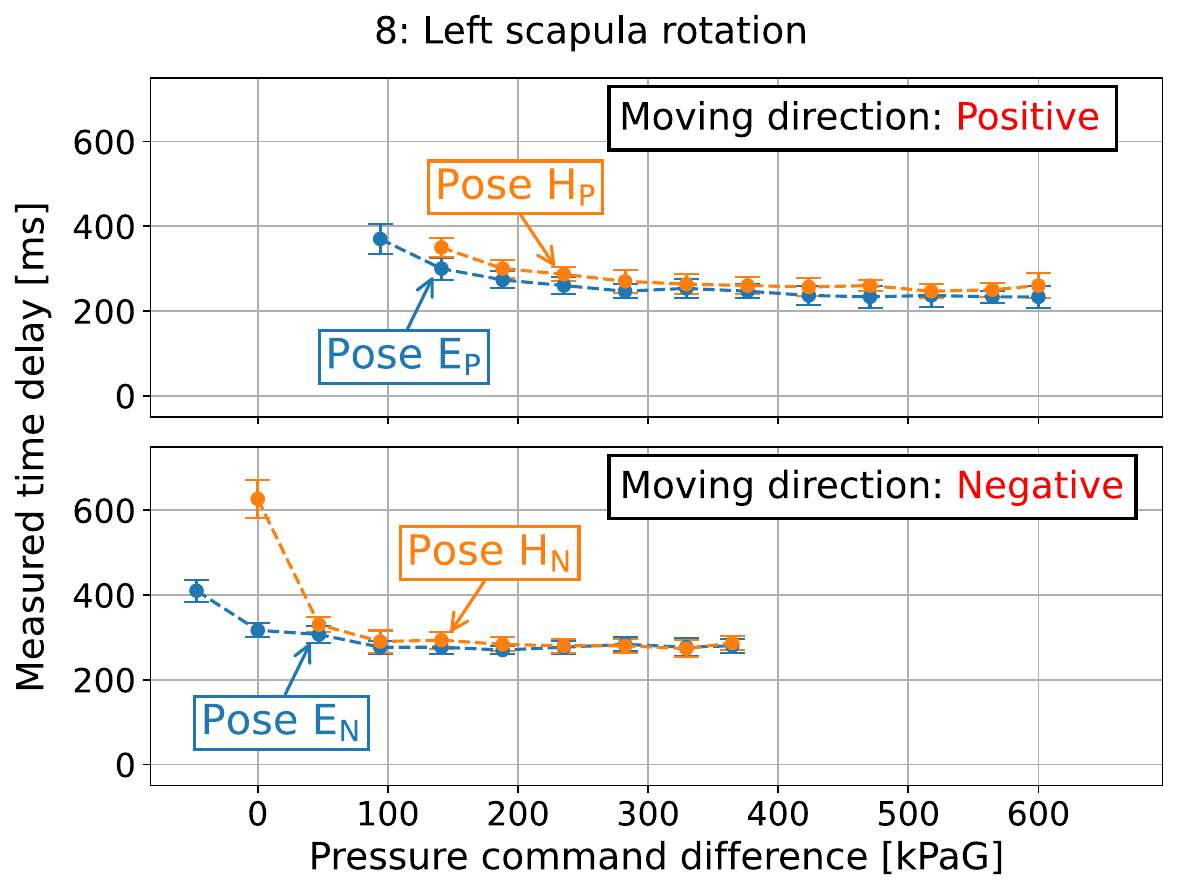}
  \caption{Estimated time delay versus applied pressure command difference for the left scapula rotation joint (joint 8). The delay converges to a stable value of approximately 280~ms for large command differences. Error bars represent standard deviation over 10 trials.}
  \label{fig:measure_time_delay/01_left_shoulder_elevation}
\end{figure}

\begin{table}[t]
  \footnotesize
  \centering
  \caption{Converged time delay values [ms] for each joint and condition.}
  \begin{tabular}{lcccc}
    \hline\hline
    & \multicolumn{2}{c}{Moving direction: Positive} & \multicolumn{2}{c}{Moving direction: Negative} \\
    \cline{2-3} \cline{4-5}
    Joint & Pose E\textsubscript{P} & Pose H\textsubscript{P} & Pose E\textsubscript{N} & Pose H\textsubscript{N} \\
    \hline
    \#1  Waist             & $ 275.56 \pm 21.12 $ & $ 274.44 \pm 18.55 $ & $ 265.56 \pm 20.73 $ & $ 266.67 \pm 19.91 $ \\
    \#8  L. Scapula        & $ 242.22 \pm 21.37 $ & $ 261.48 \pm 20.45 $ & $ 277.14 \pm 15.74 $ & $ 283.81 \pm 18.81 $ \\
    \#9  L. Shoulder Abd.  & $ 227.22 \pm 18.92 $ & $ 245.56 \pm 20.10 $ & $ 293.89 \pm 19.68 $ & $ 294.44 \pm 16.32 $ \\
    \#10 L. Shoulder Flex. & $ 254.67 \pm 19.99 $ & $ 260.33 \pm 18.58 $ & $ 285.42 \pm 20.88 $ & $ 324.44 \pm 17.10 $ \\
    \#11 L. Shoulder Rot.  & $ 235.15 \pm 18.97 $ & $ 240.61 \pm 20.35 $ & $ 228.67 \pm 16.90 $ & $ 247.33 \pm 19.27 $ \\
    \#12 L. Elbow Flex.    & $ 243.33 \pm 21.41 $ & $ 237.00 \pm 20.97 $ & $ 244.85 \pm 19.22 $ & $ 256.36 \pm 19.01 $ \\
    \#13 L. Forearm Rot.   & $ 236.67 \pm 18.16 $ & $ - $ & $ 231.52 \pm 20.79 $ & $ - $ \\
    \hline\hline
  \end{tabular}
  \label{tab:estimated_time_delay}
\end{table}

Figure~\ref{fig:measure_time_delay/01_left_shoulder_elevation} shows the measured time delay for the left scapula rotation joint (joint 8) as a function of the applied pressure command difference, where the pressure command difference is calculated as follows:
\begin{align}
  u_{\mathrm{diff},i} =
  \begin{cases}
    u^{\mathrm{A}}_{i}(t_{0}) - u^{\mathrm{B}}_{i}(0) &\text{  (Moving direction: positive)} \\
    u^{\mathrm{B}}_{i}(t_{0}) - u^{\mathrm{A}}_{i}(0) &\text{  (Moving direction: negative)}
  \end{cases}.
\end{align}
At low command differences, the time delay is long and shows significant inconsistency. However, as the command difference increases, the delay shortens and converges toward a specific value. This convergent behavior was observed across all tested joints.

Table~\ref{tab:estimated_time_delay} summarizes these converged time delay values, in which mean and standard deviation over 10 measurements are shown. These values were calculated by averaging the last several data points for each condition where the delay had saturated. The results confirm a substantial and consistent baseline delay inherent to the air transmission lines, roughly ranging from $230$~ms to $320$~ms. The posture also has a minor but consistent influence, with the hardest-to-move postures (Poses H\textsubscript{P} and H\textsubscript{N}) often resulting in a slightly longer delay than their easiest-to-move counterparts (Poses E\textsubscript{P} and E\textsubscript{N}). The low standard deviation in all cases suggests this phenomenon is highly reproducible.

\subsection{Experiment I-B: Measuring Minimum Pressure Command}
\label{sec:exp-min-pressure}

\subsubsection{Method}

\begin{figure}[t]
  \centering
  \subfloat[Positive direction]{%
    \resizebox*{0.49\linewidth}{!}{%
      \includegraphics{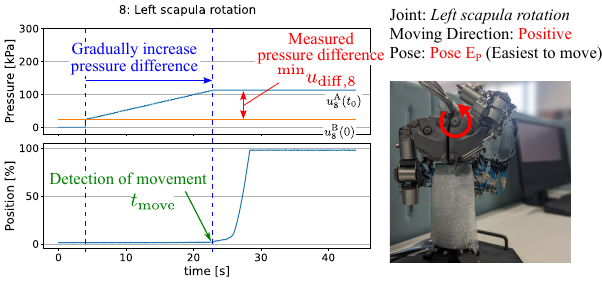}}}
  \label{fig:measure_minimum_pressure_a}
  \subfloat[Negative direction]{%
    \resizebox*{0.49\linewidth}{!}{%
      \includegraphics{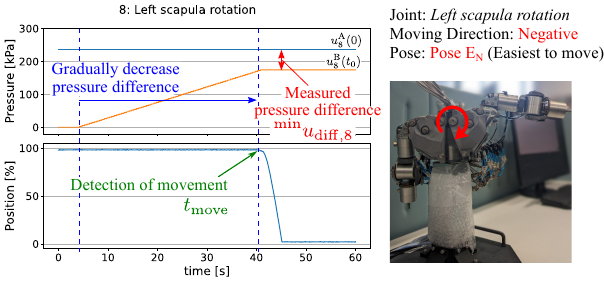}}}
  \label{fig:measure_minimum_pressure_b}
  \caption{Method for measuring the minimum pressure command difference (${}^{\mathrm{min}}u_{\mathrm{diff},8}$) for the left scapula rotation joint. The command difference is recorded at the moment movement is detected.}
  \label{fig:measure_minimum_pressure}
\end{figure}

To initiate movement, pneumatic actuators must generate sufficient force to overcome internal static friction. This experiment measures the minimum pressure command difference (${}^{\mathrm{min}}u_{\mathrm{diff},i}$) required to start movement. The procedure, illustrated in Figure~\ref{fig:measure_minimum_pressure}, was repeated 10 times for each condition.
\begin{enumerate}
\item The robot was set to an initial posture (Pose E\textsubscript{P}, H\textsubscript{P}, E\textsubscript{N} or H\textsubscript{N}) and held at a mechanical limit.
\item The pressure in one chamber was gradually increased (or decreased) while the other was held constant.
\item The command difference ${}^{\mathrm{min}}u_{\mathrm{diff},i}$ was recorded at the exact moment joint movement was detected, which was computed as follows:
\begin{align}
  {}^{\mathrm{min}}u_{\mathrm{diff},i}=
  \begin{cases}
    u^{\mathrm{A}}_{i}(t_{\mathrm{move}}) - u^{\mathrm{B}}_{i}(0) &\text{  (Moving direction: positive)} \\
    u^{\mathrm{B}}_{i}(t_{\mathrm{move}}) - u^{\mathrm{A}}_{i}(0) &\text{  (Moving direction: negative)}
  \end{cases}.
\end{align}  
\end{enumerate}






\subsubsection{Results}

\begin{figure}[t]
  \centering
  \includegraphics[width=\linewidth]{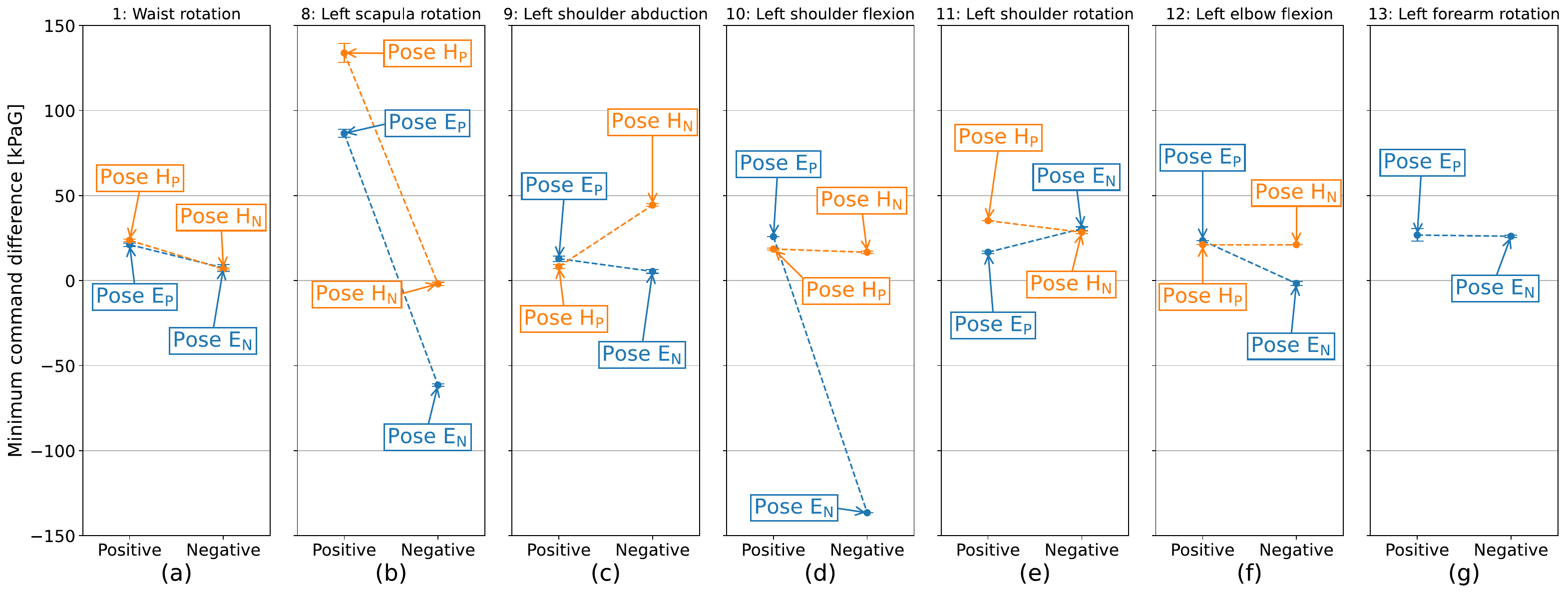}
  \caption{Average minimum pressure command difference required to initiate movement for each tested joint. The plot compares the four initial postures: Pose E\textsubscript{P} and Pose H\textsubscript{P} for the easiest and hardest positive movements, and Pose E\textsubscript{N} and Pose H\textsubscript{N} for the easiest and hardest negative movements. Error bars represent the standard deviation over 10 trials.}
  \label{fig:measure_minimum_pressure/collated_minimum_pressure_difference}
\end{figure}

The average minimum command differences for all tested joints are shown in Figure~\ref{fig:measure_minimum_pressure/collated_minimum_pressure_difference}. The results reveal that the activation threshold is affected by both intrinsic actuator properties and external loads. For joints where gravity has a negligible influence on the movement, such as waist rotation (Figure~\ref{fig:measure_minimum_pressure/collated_minimum_pressure_difference}(a)) and forearm rotation (Figure~\ref{fig:measure_minimum_pressure/collated_minimum_pressure_difference}(g)), the required command difference shows no significant variation between the easiest (Pose E\textsubscript{P} or E\textsubscript{N}) and hardest (Pose H\textsubscript{P} or H\textsubscript{N}) postures. For those joints, the activation threshold is primarily determined by the actuator's internal static friction. In contrast, for joints working against gravity, such as scapula rotation (Figure~\ref{fig:measure_minimum_pressure/collated_minimum_pressure_difference}(b)) and shoulder flexion (Figure~\ref{fig:measure_minimum_pressure/collated_minimum_pressure_difference}(d)), the external load becomes a significant factor; for positive movements against gravity, the hardest posture (Pose H\textsubscript{P}) requires a substantially higher command difference than the easiest posture (Pose E\textsubscript{P}). Across all conditions, the low variance in the measurements indicates that the static friction for each joint is consistent. While this predictability is useful, it also highlights a challenge in pneumatic control, as this friction can lead to stick-slip effects, making extremely smooth, slow motions difficult to achieve.

\subsection{Experiment I-C: Evaluating Maximum Velocity Characteristics}
\label{sec:exp-max-velocity}

\subsubsection{Method}

To understand the robot's capacity for dynamic movement, this experiment measured the maximum joint velocity (${}^{\mathrm{max}}v_{i}$) under the maximum possible pressure difference. The procedure was repeated 10 times for each condition.
\begin{enumerate}
    \item The robot was set to an initial posture (Pose E\textsubscript{P}, H\textsubscript{P}, E\textsubscript{N} or H\textsubscript{N}) and held at a mechanical limit.
    \item At $t_{0}$, a maximum step input was applied by setting $u^{\mathrm{A}}_{i}$ to max.
    \item The resulting joint trajectory $q_i(t)$ was recorded, and the maximum velocity was calculated from its numerical time derivative.
\end{enumerate}

\subsubsection{Results}

\begin{figure}[t]
  \centering
  \subfloat[Positive direction]{%
    \resizebox*{0.49\linewidth}{!}{%
      \includegraphics{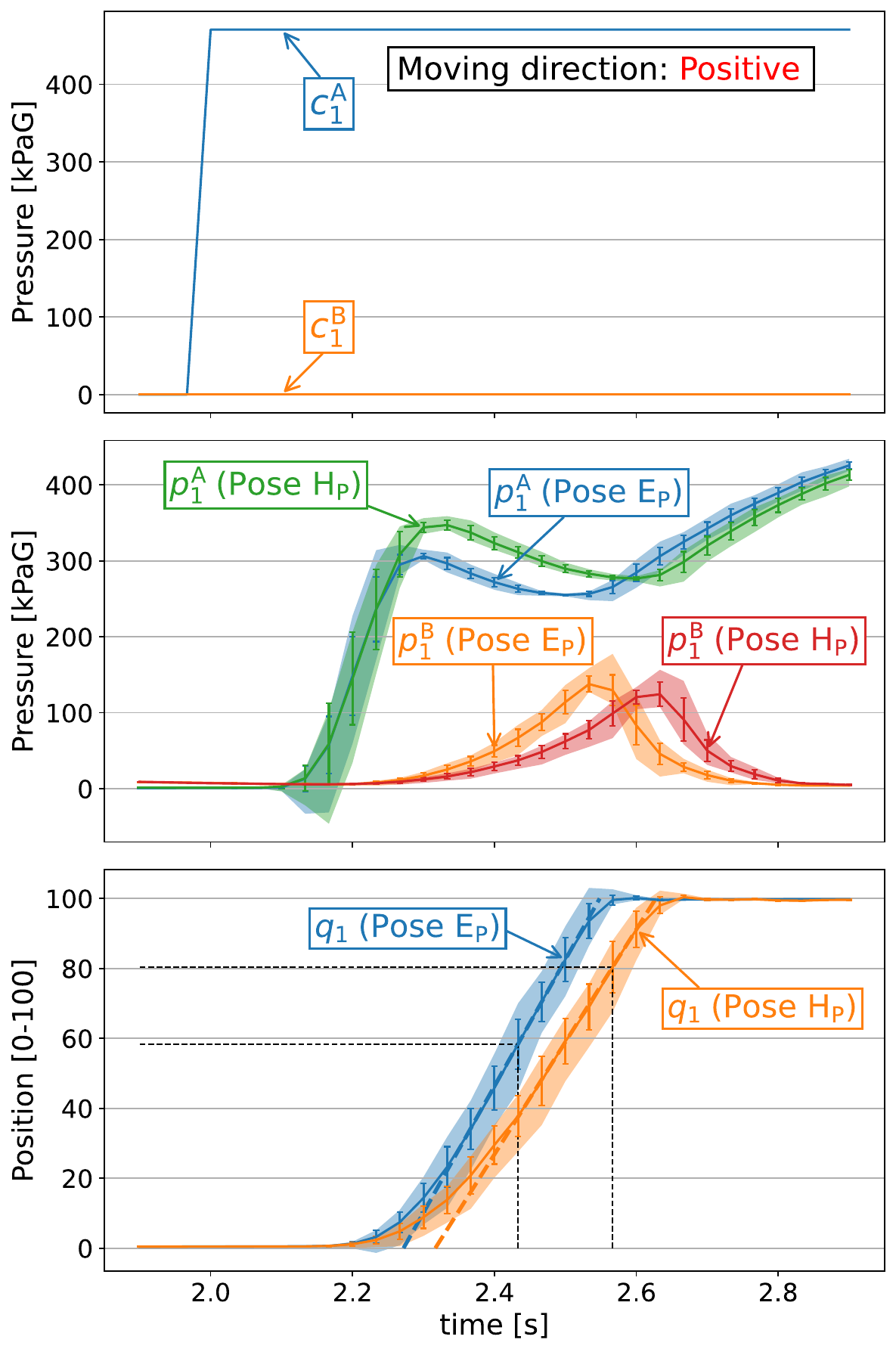}}}
  \label{fig:observe_max_pressure/positive/00_waist}
  \subfloat[Negative direction]{%
    \resizebox*{0.49\linewidth}{!}{%
      \includegraphics{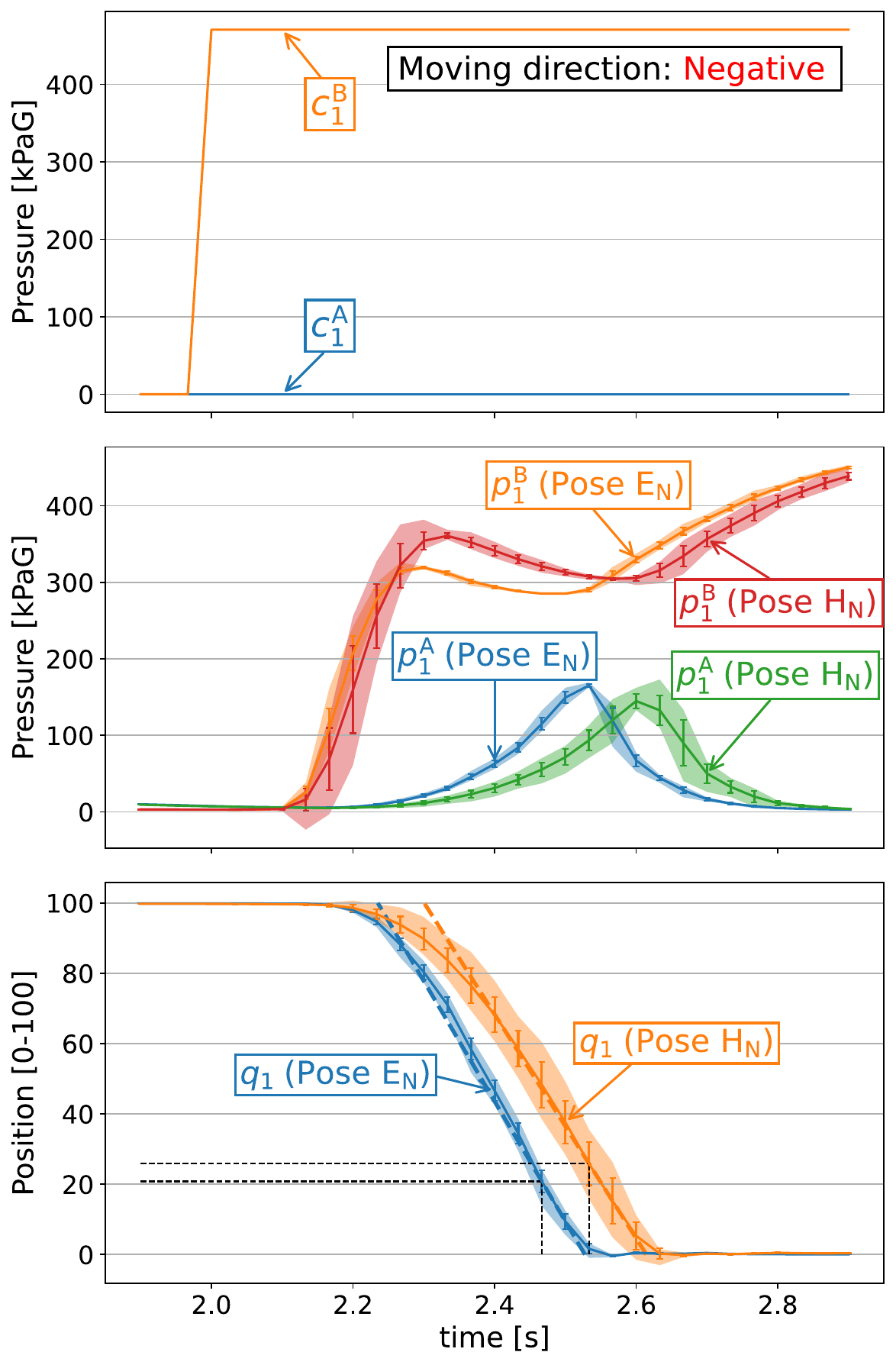}}}
  \label{fig:observe_max_pressure/negative/00_waist}
  \caption{Motion profiles for the waist joint (joint 1) under maximum pressure command for (a) positive and (b) negative movement directions. The top, middle, and bottom plots show the commanded pressure, measured pressure responses, and resulting joint trajectories, respectively. The trajectories for the easiest (Pose E\textsubscript{P} in (a), Pose E\textsubscript{N} in (b)) and hardest (Pose H\textsubscript{P} in (a), Pose H\textsubscript{N} in (b)) postures are compared, showing a clear difference in speed.}
  \label{fig:observe_max_pressure/00_waist}
\end{figure}

\begin{figure}[t]
  \centering
  \subfloat[Positive direction]{%
    \resizebox*{0.49\linewidth}{!}{%
      \includegraphics{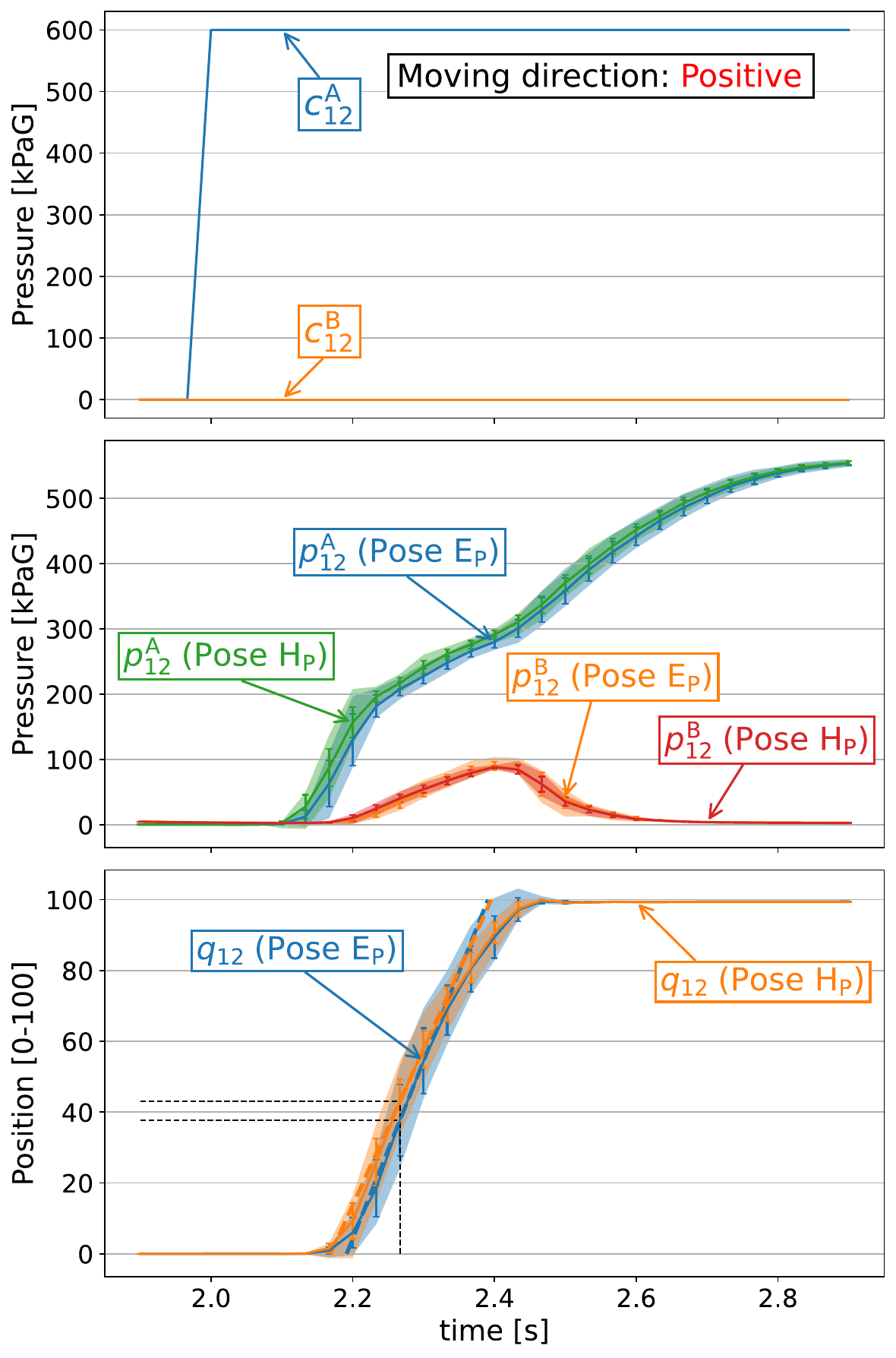}}}
  \label{fig:observe_max_pressure/positive/05_left_elbow_flexion}
  \subfloat[Negative direction]{%
    \resizebox*{0.49\linewidth}{!}{%
      \includegraphics{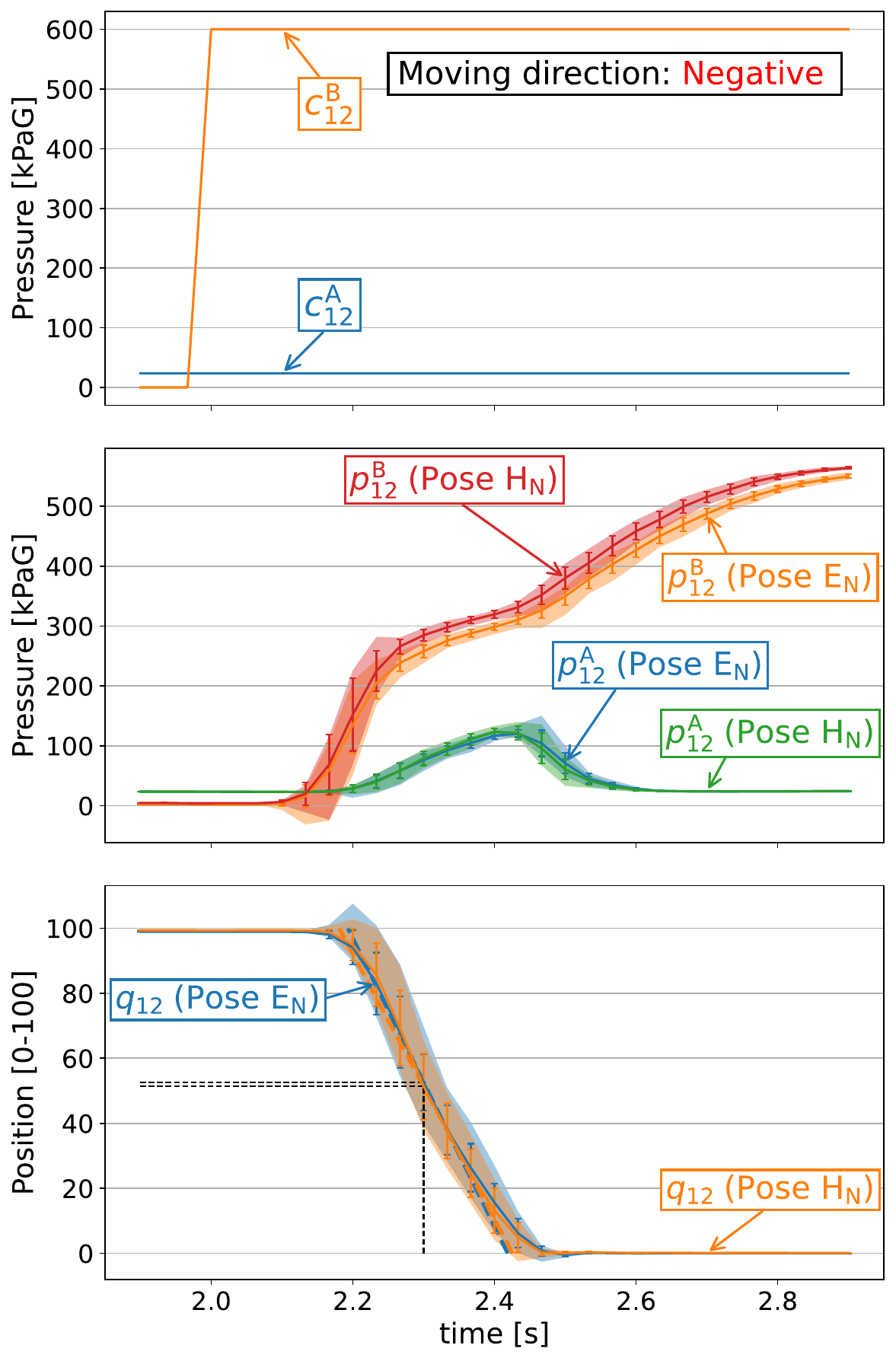}}}
  \label{fig:observe_max_pressure/negative/05_left_elbow_flexion}
  \caption{Motion profiles for the left elbow flexion joint (joint 12) under maximum pressure command for (a) positive and (b) negative movement directions. The plot layout is identical to that of Figure~\ref{fig:observe_max_pressure/00_waist}. The trajectories for the easiest (Pose E\textsubscript{P} in (a), Pose E\textsubscript{N} in (b)) and hardest (Pose H\textsubscript{P} in (a), Pose H\textsubscript{N} in (b)) postures are compared. For this distal joint, the influence of posture on the resulting trajectories is less significant compared to the waist joint. }
  \label{fig:observe_max_pressure/05_left_elbow_flexion}
\end{figure}

\begin{table}[t]
  \footnotesize
  \centering
  \caption{Maximum joint velocity [0-100/s] for each joint and condition.}
  \begin{tabular}{lcccc}
    \hline\hline
    & \multicolumn{2}{c}{Moving direction: Positive} & \multicolumn{2}{c}{Moving direction: Negative} \\
    \cline{2-3} \cline{4-5}
    Joint No. & Pose E\textsubscript{P} & Pose H\textsubscript{P} & Pose E\textsubscript{N} & Pose H\textsubscript{N} \\
    \hline
    \#1  Waist             & $ 445.47 \pm 8.36 $ & $ 394.42 \pm 8.81 $ & $ -451.82 \pm 6.85 $ & $ -401.07 \pm 7.25 $ \\
    \#8  L. Scapula        & $ 816.20 \pm 98.93 $ & $ 797.04 \pm 51.17 $ & $ -442.74 \pm 48.88 $ & $ -379.94 \pm 32.68 $ \\
    \#9  L. Shoulder Abd.  & $ 395.31 \pm 22.32 $ & $ 439.86 \pm 58.97 $ & $ -222.81 \pm 28.19 $ & $ -182.50 \pm 19.40 $ \\
    \#10 L. Shoulder Flex. & $ 728.07 \pm 52.06 $ & $ 559.97 \pm 17.96 $ & $ -415.80 \pm 21.10 $ & $ -328.21 \pm 7.84 $ \\
    \#11 L. Shoulder Rot.  & $ 641.15 \pm 68.60 $ & $ 615.60 \pm 40.99 $ & $ -572.96 \pm 40.17 $ & $ -638.92 \pm 59.21 $ \\
    \#12 L. Elbow Flex.    & $ 626.83 \pm 54.45 $ & $ 564.33 \pm 61.26 $ & $ -555.74 \pm 48.20 $ & $ -588.55 \pm 66.67 $ \\
    \#13 L. Forearm Rot.   & $ 613.95 \pm 58.85 $ & $ - $ & $ -552.82 \pm 25.16 $ & $ - $ \\
    \hline\hline
  \end{tabular}
  \label{tab:calculated_max_velocity}
\end{table}

The resulting motion profiles for the waist and left elbow flexion joints are shown in Figures~\ref{fig:observe_max_pressure/00_waist} and \ref{fig:observe_max_pressure/05_left_elbow_flexion}, and the maximum velocities for all joints are summarized in Table~\ref{tab:calculated_max_velocity}. For each joint, the figures display the commanded pressure step input (top plot), the measured pressure responses in each actuator chamber (middle plot), and the resulting joint angle trajectories (bottom plot). The key finding is that the robot's posture has a much larger impact on the dynamic performance of proximal joints (close to the torso) than distal joints.

As seen in Figure~\ref{fig:observe_max_pressure/00_waist}, the motion profile of the waist joint differs significantly between the easiest (Pose E\textsubscript{P} or E\textsubscript{N}) and hardest (Pose H\textsubscript{P} or H\textsubscript{N}) postures, as its movement is affected by the inertia of the entire upper body. In contrast, for a distal joint like the elbow (Figure~\ref{fig:observe_max_pressure/05_left_elbow_flexion}), the difference between these two postural conditions is much smaller. This configuration-dependent behavior underscores the difficulty of developing a single, simple dynamic model for the robot.

\subsection{Experiment II: Assessing Motion Reproducibility}
\label{sec:exp-reproducibility}

\subsubsection{Method}

The previous experiments suggest that while the robot's dynamics are highly complex and nonlinear, they are also consistent. This final experiment aims to quantify this consistency, or reproducibility, which is a prerequisite for data-driven control.
\begin{enumerate}
\item The robot was initialized to a consistent starting state by moving all joints to one of their mechanical limits.
\item A predetermined, random time series of pressure commands was sent to all 26 valves simultaneously for 60 seconds.
\item The resulting joint angle trajectories $\bm{q}(t)$ were recorded.
\item The procedure was repeated 11 times. The first trial was discarded as a warm-up.
\item The similarity between any two trial trajectories ($j$ and $k$) for a single joint $i$ was quantified by calculating the Root Mean Square Error (RMSE):
  \begin{align}
    \mathrm{RMSE}_{i}(j, k) = \sqrt{\frac{1}{T}\sum_{t=1}^{T}(q_{i}^{(j)}(t) - q_{i}^{(k)}(t))^2}.
  \end{align}
\end{enumerate}

\subsubsection{Results}

\begin{figure}
  \centering
  \subfloat[Waist joint (1)]{%
    \resizebox*{0.3\linewidth}{!}{%
      \includegraphics{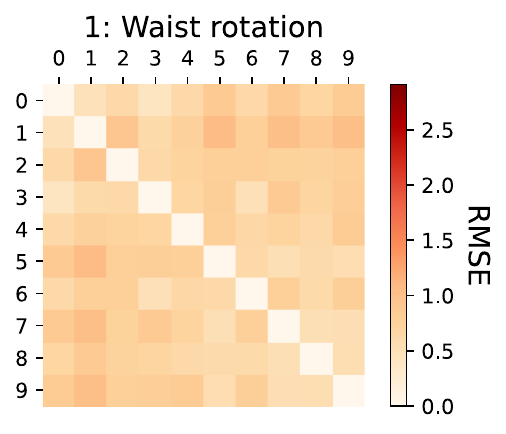}}}
  \label{fig:reproduce_motion/rmse_0}
  \hfill
  \subfloat[Arm joints (2-7, 8-13)]{%
    \resizebox*{0.65\linewidth}{!}{%
      \includegraphics[width=0.65\linewidth]{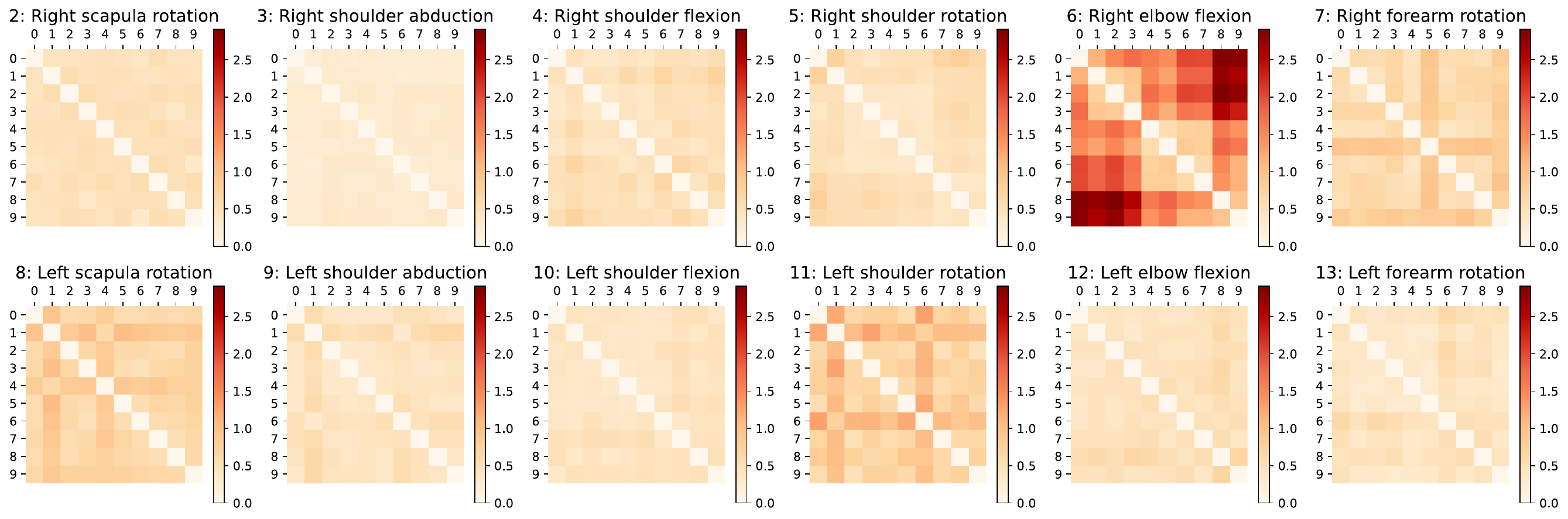}}}
  \label{fig:reproduce_motion/rmse_1-12}
  \caption{RMSE matrices evaluating the trial-to-trial reproducibility for (a) the waist joint and (b) the 12 arm joints. Each cell $(j,\;k)$ shows the RMSE between trial $j$ and trial $k$. The light coloring indicates very high similarity across trials.}
  \label{fig:reproduce_motion/rmse}
\end{figure}

\begin{figure}[t]
  \centering
  \includegraphics[width=\linewidth]{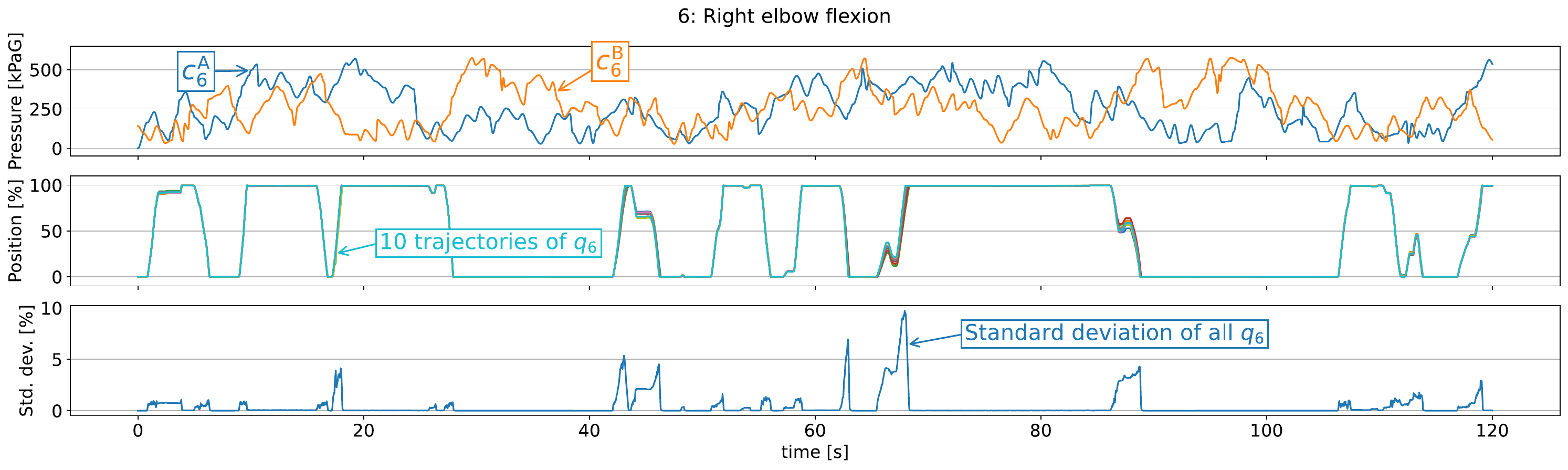}
  \caption{Reproducibility results for the right elbow flexion joint (joint 6), which exhibited the highest error. Top: random input commands. Middle: 10 overlaid trajectories showing high consistency. Bottom: standard deviation over time, which remains small.}
  \label{fig:reproduce_motion/11_right_elbow_flexion}
\end{figure}

The RMSE values for all pairs of trials were computed for each joint and are visualized as matrices in Figure~\ref{fig:reproduce_motion/rmse}. The color in each cell represents the RMSE between two trials; lighter colors indicate higher similarity (lower error). The matrices are overwhelmingly light, indicating a very low RMSE between trials for almost all joints. The largest errors were observed in the right elbow flexion joint (joint 6). Because joints 6 and 12 are structurally symmetric, the difference is unlikely to arise from the nominal joint design itself. One plausible explanation is actuator-to-actuator variability in internal friction between the two customized actuators. Since pneumatic joint motion is sensitive to such friction differences, even small variations can appear as minor differences in motion reproducibility. This explanation has not been tested experimentally, and confirming it would require friction measurements on the individual actuators. However, a closer look at Figure~\ref{fig:reproduce_motion/11_right_elbow_flexion} reveals that while the middle plot shows some small trial-to-trial variations in the trajectories, the bottom plot confirms these are minor relative to the joint's overall range of motion by illustrating the consistently low standard deviation.

The results indicate that the robot's behavior is highly reproducible under the tested conditions. Given the same initial state and the same command sequence, the system follows a nearly identical trajectory. This deterministic nature, despite the underlying dynamic complexity, supports the feasibility of a data-driven approach for precise motion control.

\section{Data-Driven Trajectory Tracking Control}
\label{sec:data-driven-control}

The experiments in Section~\ref{sec:dynamic-properties} demonstrated that the robot, while exhibiting highly reproducible behavior, possesses complex, nonlinear dynamics and a significant actuation delay. These characteristics make it difficult to control accurately using traditional model-based controllers. This section presents a feasibility study on a data-driven controller designed to overcome these challenges. The approach involves two stages: (1) collecting a comprehensive dataset of the robot's dynamic behavior, and (2) using this data to train an inverse dynamics model that explicitly compensates for the system's time delay.

\subsection{Motion Data Collection}
\label{sec:data-collection}

\begin{figure}[t]
  \centering
  \includegraphics[width=0.5\linewidth]{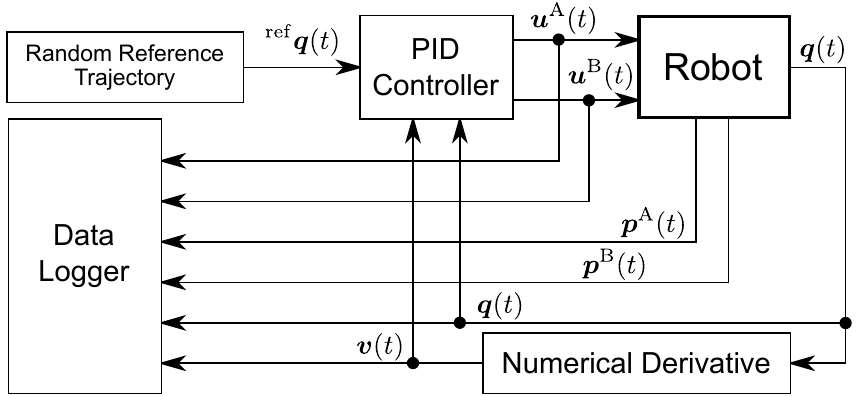}
  \caption{Block diagram of the data collection framework. A PID controller drives the robot to follow a randomly generated trajectory, while a data logger records the resulting time-series data of commands and sensor values.}
  \label{fig:block_diagram_data_collection}
\end{figure}

\begin{figure}[t]
  \centering
  \includegraphics[width=\linewidth]{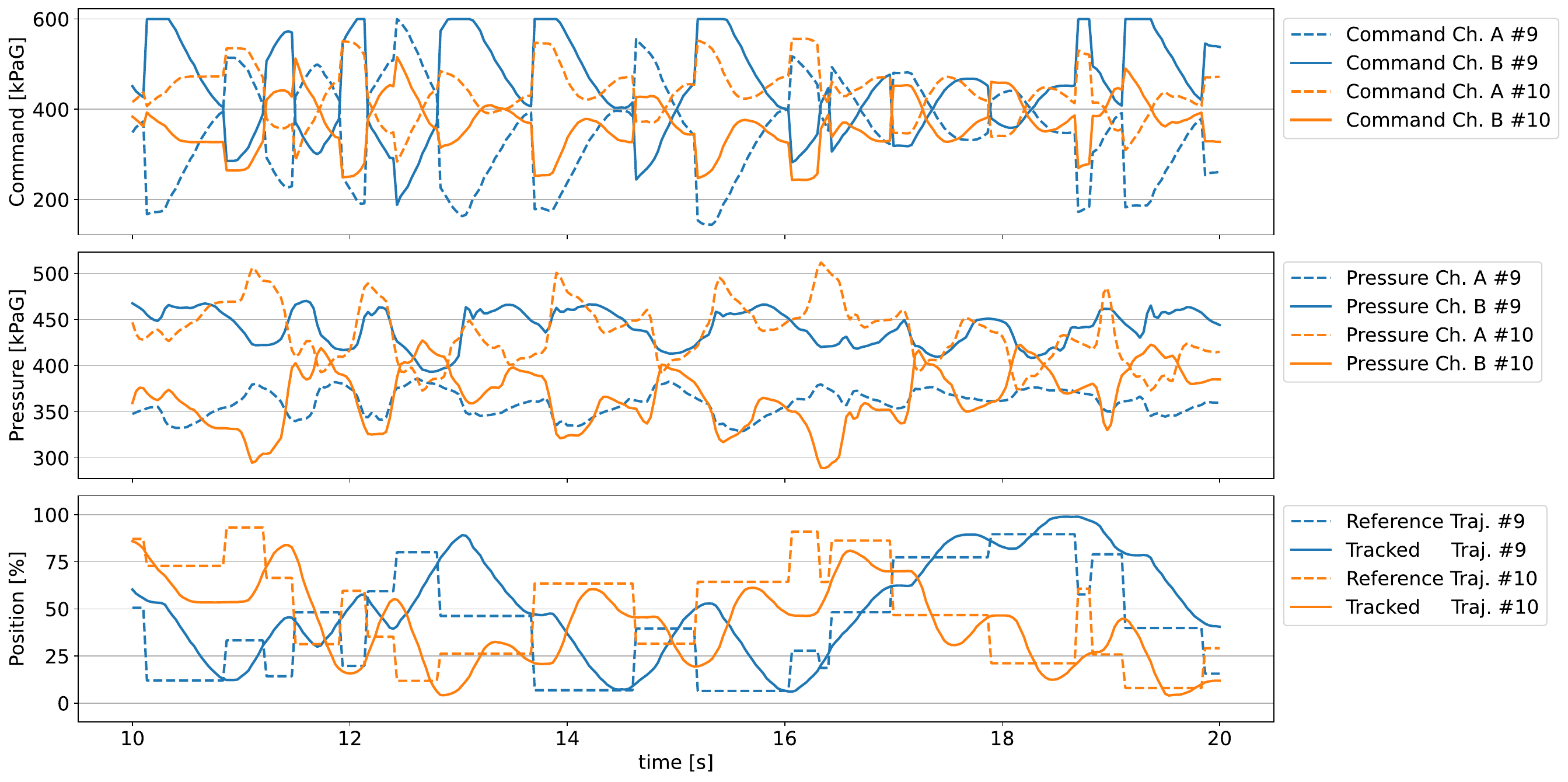}
  \caption{An example of the collected time-series data for joint \#9 (blue) and joint \#10 (orange). The top plot shows the valve commands computed by the PID controller for Chamber A (solid lines) and Chamber B (dashed lines). These commands produce the measured chamber pressures shown in the middle plot. The bottom plot displays the randomly generated reference trajectories the controller was tasked to follow (dashed lines) and the robot's actual tracked trajectories that resulted from the applied pressures (solid lines).}
  \label{fig:data_collection_joint_2_3}
\end{figure}

The first step is to create a rich dataset that captures the relationship between control inputs (valve commands) and system states (joint angles, velocities, and pressures) across a wide range of movements. Although the motion reproducibility experiment in Section~\ref{sec:exp-reproducibility} demonstrated reproducible behavior across all 13 joints, we focused on an arm subsystem in this preliminary controller study to keep the learning problem tractable and to evaluate the time-delay compensation strategy under coupled multi-joint motion. This subsystem was chosen for its functional significance and dynamic challenges. Functionally, the selected joints are the primary actuators responsible for positioning the arm in its workspace for representative tasks like reaching. Dynamically, they form a highly coupled system that includes joints heavily affected by gravity. Based on these criteria, we selected the 4-DOF subsystem comprising the left arm's shoulder abduction (joint 9), shoulder flexion (joint 10), shoulder rotation (joint 11), and elbow flexion (joint 12).

We identified the 4-DOF arm simultaneously so that the model could learn the dynamic coupling between joints directly from data. Even if single-DOF models were learned separately, coordinated 4-DOF motion would still require additional treatment of interaction effects among joints. These effects are not the rigid-body terms alone but load-dependent changes in the friction, delay, and stiffness of each actuator, as explained in Section~\ref{sec:interaction-effects}. In addition, each joint has different actuator and transmission-tube characteristics.

\subsubsection{Data Generation Method}

To ensure the data contained sufficient dynamic richness for training, we used a simple PID controller to make the robot follow a randomly generated reference trajectory, as shown in the block diagram in Figure~\ref{fig:block_diagram_data_collection}. It is important to note that the PID controller's purpose was not to achieve perfect tracking, but merely to serve as a driver to generate smooth, continuous, and varied movements that explore the system's state space.

The random reference trajectory ${}^{\mathrm{ref}}\bm{q}(t)$ was generated using a random walk, where the target angle for each joint was updated by a random increment at intervals chosen randomly between $0.1$ and $1.0$~s. The PID control law is given by:
\begin{align}
  \bm{e}(t) &= {}^{\mathrm{ref}}\bm{q}(t) - \bm{q}(t) \\
  \Delta \bm{u}(t) &= \bm{K}_{\mathrm{p}}\bm{e}(t) + \bm{K}_{\mathrm{d}}\frac{d\bm{e}(t)}{dt} + \bm{K}_{\mathrm{i}}\int_{0}^{t}\bm{e}(\tau)d\tau \label{eq:pid_control} \\
  \bm{u}^{\mathrm{A}}(t) &= \bm{u}_{0} + \frac{1}{2}\Delta \bm{u}(t) \quad , \quad \bm{u}^{\mathrm{B}}(t) = \bm{u}_{0} - \frac{1}{2}\Delta \bm{u}(t),
\end{align}
where $\bm{K}_{\mathrm{p, d, i}}$ are diagonal gain matrices and $\bm{u}_0$ is a baseline pressure command vector that defines the joint's passive stiffness. The manually tuned gains and baseline stiffness were set to $\bm{K}_{\mathrm{p}}=\mathrm{diag}\{5.48, 3.08, 3.64, 3.4\}$, $\bm{K}_{\mathrm{d}}=\mathrm{diag}\{0.044, 0.148, 0.016, 0.016\}$, $\bm{K}_{\mathrm{i}}=\mathrm{diag}\{0.0005, 0.0036, 0.001, 0.0009\}$, and $\bm{u}_{0} = [400.0, 400.0, 400.0, 400.0]$. Figure~\ref{fig:data_collection_joint_2_3} shows a sample of the rich time-series data generated with this method.

\subsubsection{Collected Dataset}

The robot performed 100 trials, each lasting 60 seconds. During these trials, the valve commands ($\bm{u}^{\mathrm{A}}$, $\bm{u}^{\mathrm{B}}$), chamber pressures ($\bm{p}^{\mathrm{A}}$, $\bm{p}^{\mathrm{B}}$), joint angles ($\bm{q}$), and joint velocities ($\bm{v}$) were all recorded at $30$~Hz. This process yielded a final dataset containing 180,000 time-stamped data points for the four selected joints. The robot operated continuously without malfunction, confirming its mechanical robustness for long-duration experiments of the developed robot.

\subsection{Inverse Dynamics Model with Time-Delay Compensation}
\label{sec:learning-framework}

\begin{figure}[t]
  \centering
  \includegraphics[width=0.75\linewidth]{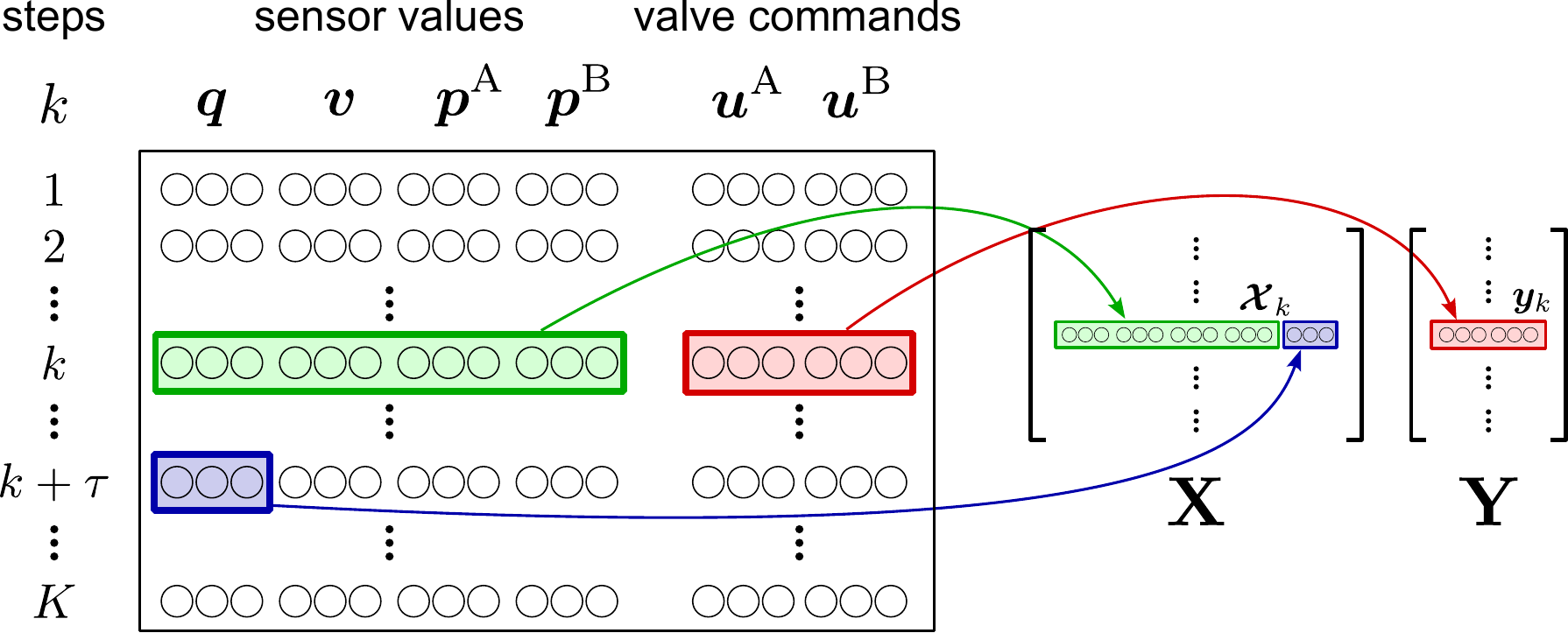}
  \caption{Schematic of the data preprocessing to compensate for time delay. An input sample $\bm{\mathcal{X}}_{k}$ is formed by concatenating current sensor values with desired future angles from $\tau$ steps ahead, while the corresponding output $\bm{y}_{k}$ is the current valve command. All such samples are then stacked vertically as rows to create the final training matrices, $\mathbf{X}$ and $\mathbf{Y}$.}
  \label{fig:data_adapter_preview}
\end{figure}

\subsubsection{Time-Delay Compensation via Data Preprocessing}

The goal of the controller is to find a function that maps a desired motion to the valve commands required to achieve that motion. We frame this as learning an inverse dynamics model, $\bm{f}: \bm{\mathcal{X}}_{k} \rightarrow \bm{y}_{k}$, where $\bm{\mathcal{X}}_{k}$ represents the robot's current and desired future states and $\bm{y}_{k}$ is the corresponding actuation command vector needed to produce that outcome. As discussed in Section~\ref{sec:interaction-effects}, the rigid-body gravitational and inertial loads are, in principle, computable, but accurate control of this robot requires the full mapping from desired motion to valve commands, which also includes actuator nonlinearities, transmission delay, friction, pressure dynamics, and the load-dependent coupling effects among joints. In particular, this mapping is affected by actuator internal friction, which has been modeled for pneumatic actuators in prior studies \cite{khayati.etal2009,schluter.perondi2018}, and by the attenuation and delay introduced by the long transmission tubes, which have also been studied in the literature \cite{krichel.sawodny2014,kern2017}. However, identifying a sufficiently accurate control-oriented inverse model for the installed multi-DOF robot remains difficult in practice. Therefore, we employ a data-driven approach to learn this mapping directly from measured data in this preliminary study.

A critical challenge is the actuation delay identified in Section~\ref{sec:exp-time-delay}. To compensate for this delay, we adopt a preview-inspired strategy in which future desired states are explicitly embedded in the input feature vector during data preprocessing. This idea is related to preview control \cite{birla.swarup2015,liao.liao2018}, but it is implemented here in a data-driven framework rather than in a model-based controller. Recent data-driven studies on delay systems have mainly considered linear systems, including Smith-Predictor-based controller design for known delay and state-feedback design for linear systems with uncertain or unknown delay \cite{formentin.etal2012,rueda-escobedo.etal2022,portilla.etal2025}. These studies derive controllers through explicit linear-system synthesis formulas, whereas our method learns a nonlinear inverse mapping with future reference information embedded directly in the input features. Previous studies on long-line pneumatic systems have mainly addressed delay using model-based methods on 1-DOF setups \cite{yang.etal2011,turkseven.ueda2018,butt.sepehri2019}. In contrast, our study examines this strategy on a coupled multi-DOF pneumatic subsystem of a compact humanoid robot. As illustrated in Figure~\ref{fig:data_adapter_preview}, the input and output vectors for a single training sample at time $k$ are defined as:
\begin{align}
\text{Input:}\quad \bm{\mathcal{X}}_{k} &= [\bm{q}(k); \bm{v}(k); \bm{p}^{\mathrm{A}}(k); \bm{p}^{\mathrm{B}}(k); \bm{q}(k+\tau)] \\
\text{Output:}\quad \bm{y}_{k} &= [\bm{u}^{\mathrm{A}}(k); \bm{u}^{\mathrm{B}}(k)]
\end{align}
For the 4-DOF subsystem under study, the input vector $\bm{\mathcal{X}}_{k}$ has a dimension of 20 (a concatenation of five 4-dimensional vectors), and the output vector $\bm{y}_{k}$ has a dimension of 8 (a concatenation of two 4-dimensional vectors). From a total of $K$ recorded time steps, we construct the full training dataset, consisting of an input matrix $\mathbf{X}\in \mathbb{R}^{(K-\tau)\times 20}$ and an output matrix $\mathbf{Y}\in \mathbb{R}^{(K-\tau)\times 8}$. This formulation forces the model to learn the valve commands at the present time $k$ that will cause the robot to achieve the desired state at a future time $k+\tau$. Based on our empirical results, the lookahead period $\tau$ was set to 9 time steps ($9 \times 33.3\text{ms} \approx 300\text{ms}$).

\subsubsection{Model Architecture and Training}
A standard three-layer multilayer perceptron (MLP) was used as the function approximator for the inverse model. The network architecture included an input layer with a size of 20 to match the dimension of $\bm{\mathcal{X}}_{k}$, a hidden layer with 200 neurons using a hyperbolic tangent ($\tanh$) activation function, and an output layer with a size of 8 to match the dimension of $\bm{y}_{k}$. Before training, both the input feature matrix $\mathbf{X}$ and the output matrix $\mathbf{Y}$ were standardized by removing the mean of each column and scaling it to unit variance.

\subsection{Evaluation of Tracking Performance}
\label{sec:evaluation}

\begin{figure}[t]
  \centering
  \includegraphics[width=0.7\linewidth]{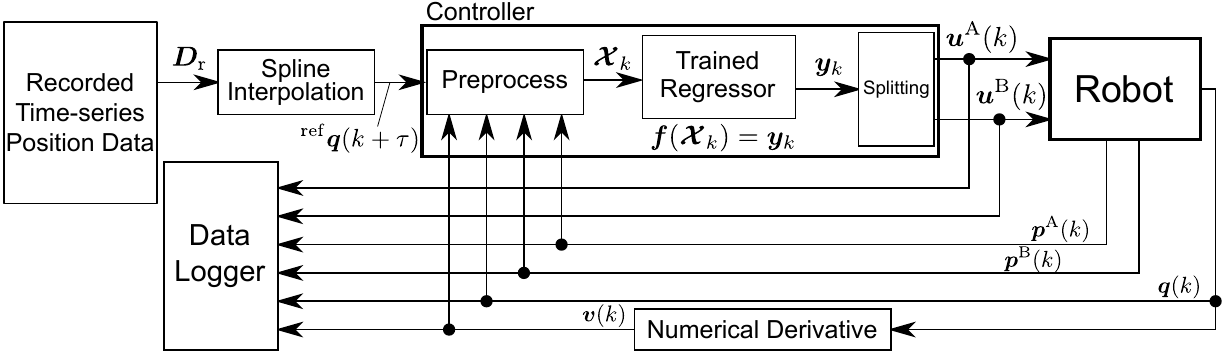}
  \caption{Control framework for trajectory tracking using the trained inverse dynamics model. The model acts as a feedback controller, taking current state and future desired angles to generate valve commands.}
  \label{fig:block_diagram_tracking}
\end{figure}

\subsubsection{Method}
To evaluate the trained model, we tasked it with tracking a new, unseen reference trajectory that was recorded by a human manually guiding the robot's arm. This trajectory was not part of the training set. The control framework is shown in Figure~\ref{fig:block_diagram_tracking}. For any given recorded trajectory $\bm{D}_{r}$, spline interpolation is used to generate a smooth, continuous desired trajectory ${}^{\mathrm{ref}}\bm{q}(t)$. At each time step $k$, the current sensor readings and the future desired angle ${}^{\mathrm{ref}}\bm{q}(k+\tau)$ are fed into the trained MLP model, which outputs the valve commands $\bm{y}_{k}=[\bm{u}^{\mathrm{A}}(k);\;\bm{u}^{\mathrm{B}}(k)]$.

The performance of this data-driven controller was compared against the same PID controller used for data collection. Each controller attempted to track the same reference trajectory 10 times.

\subsubsection{Results}

\begin{figure}[t]
  \centering
  \subfloat[Left shoulder abduction]{%
    \resizebox*{0.49\linewidth}{!}{%
      \includegraphics{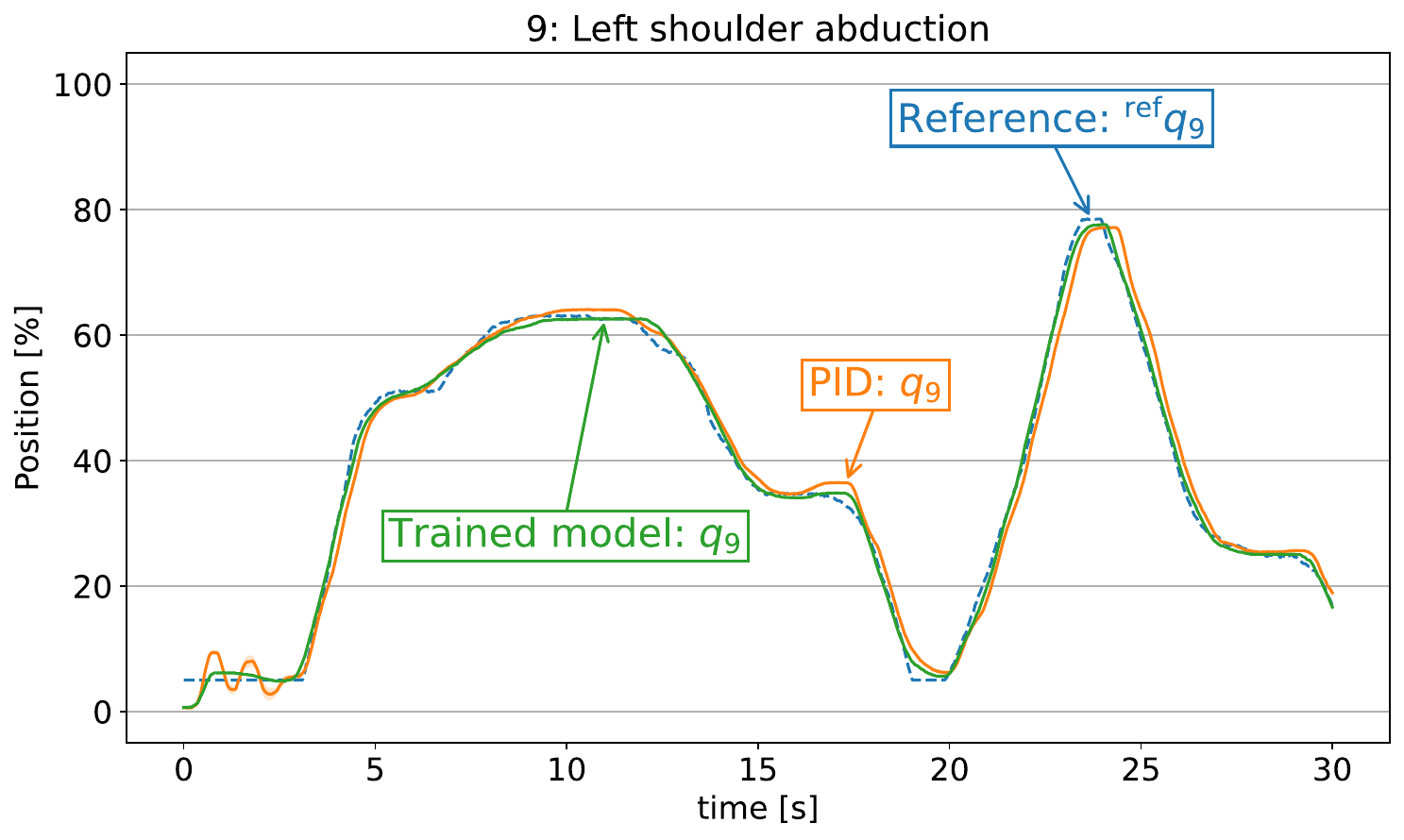}}}
  \label{fig:data_driven_ctrl/compare_trajectory_02_q}
  \subfloat[Left shoulder flexion]{%
    \resizebox*{0.49\linewidth}{!}{%
      \includegraphics{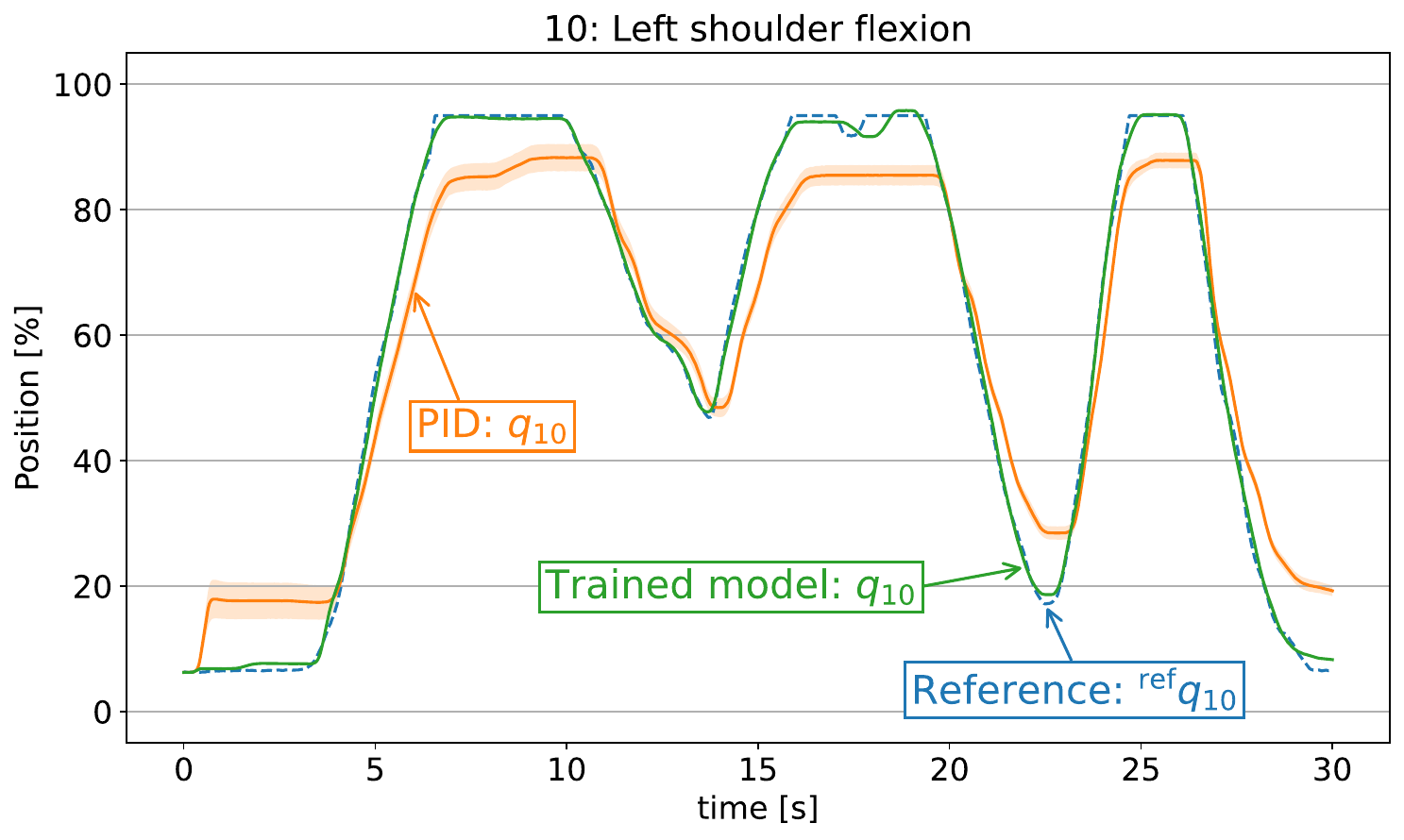}}}
  \label{fig:data_driven_ctrl/compare_trajectory_03_q}\\
  \subfloat[Left shoulder rotation]{%
    \resizebox*{0.49\linewidth}{!}{%
      \includegraphics{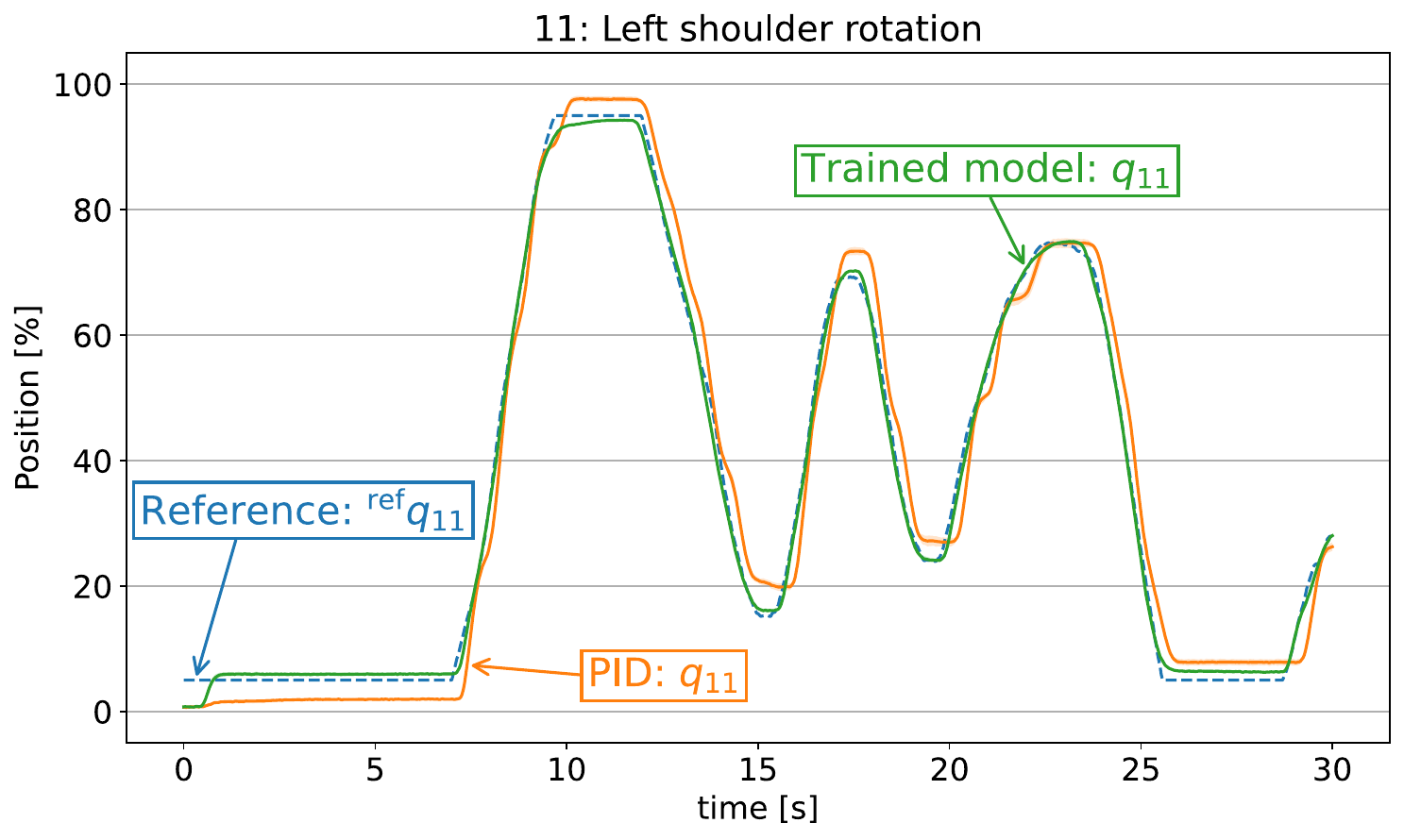}}}
  \label{fig:data_driven_ctrl/compare_trajectory_04_q}
  \subfloat[Left elbow flexion]{%
    \resizebox*{0.49\linewidth}{!}{%
      \includegraphics{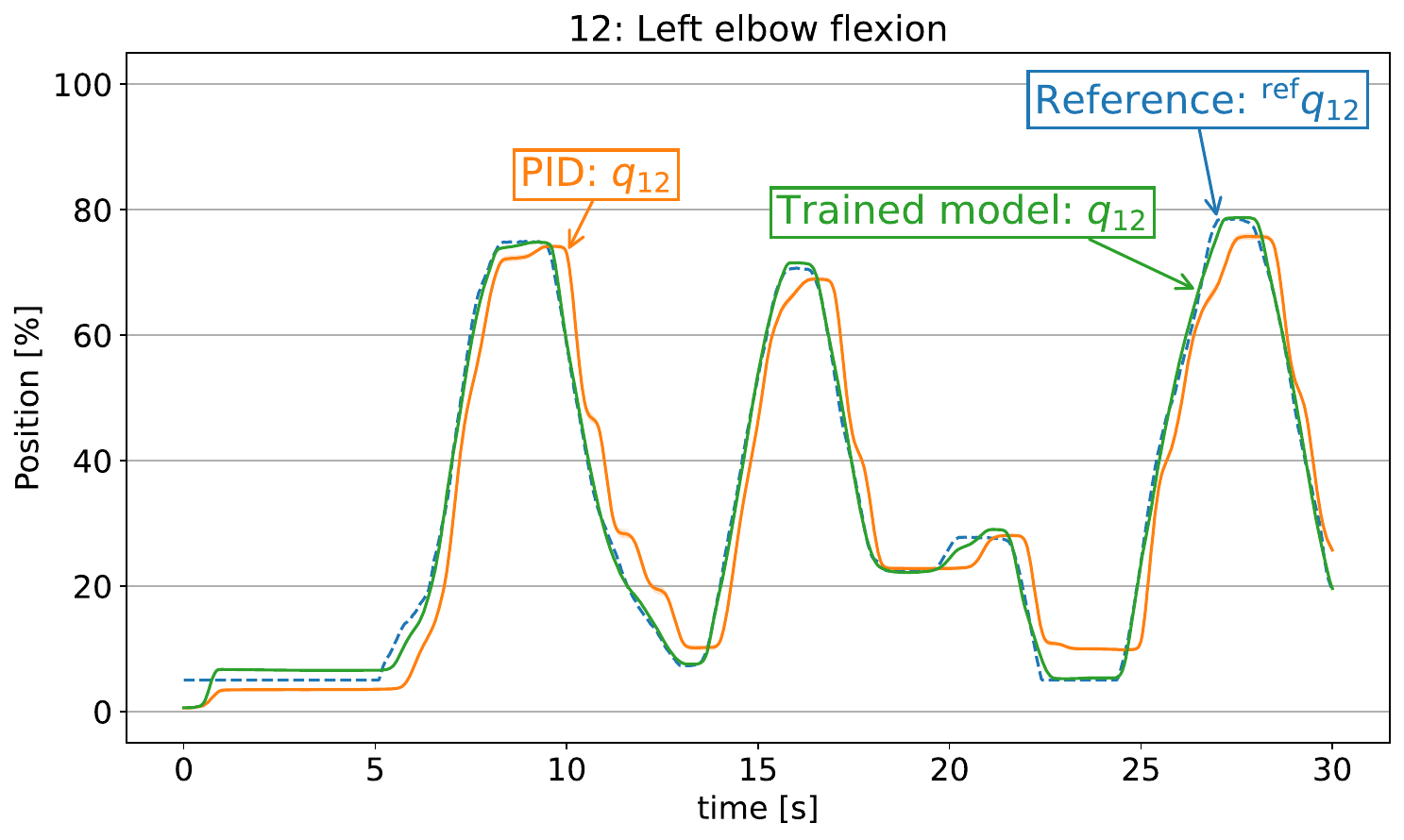}}}
  \label{fig:data_driven_ctrl/compare_trajectory_05_q}
  \caption{Comparison of trajectory tracking performance for four left arm joints. The trained model (green) follows the reference trajectory (dashed blue) much more closely than the PID controller (orange), which exhibits significant lag and error. Shaded areas represent standard deviation over 10 trials.}
  \label{fig:data_driven_ctrl/compare_trajectory}
\end{figure}

\begin{figure}[t]
  \centering
  \includegraphics[width=.7\linewidth]{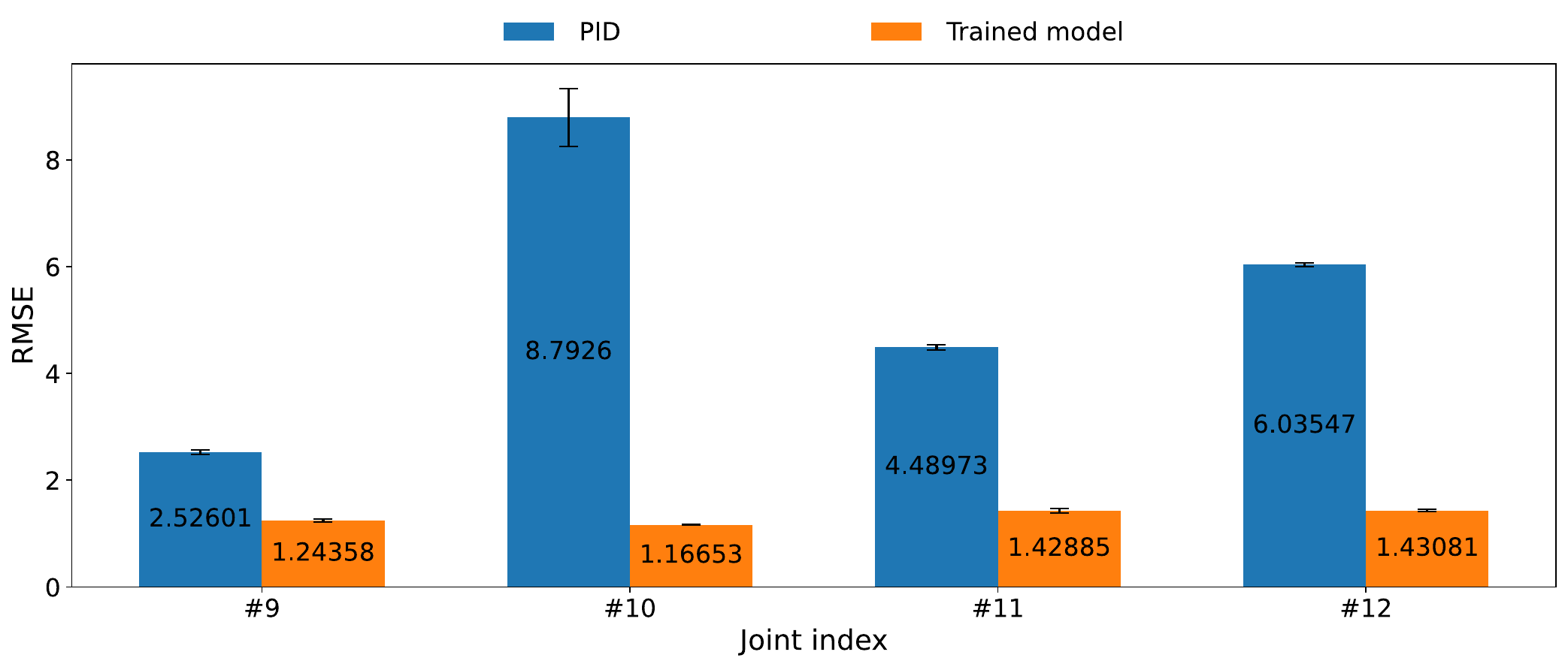}
  \caption{Comparison of RMSE for the PID and the trained model controllers. The data-driven model achieves lower tracking error across all tested joints.}
  \label{fig:data_driven_ctrl/compare_rmse}
\end{figure}

The trajectory tracking results are shown in Figure~\ref{fig:data_driven_ctrl/compare_trajectory}, and the quantitative performance is summarized by the Root Mean Square Error (RMSE) in Figure~\ref{fig:data_driven_ctrl/compare_rmse}.
The results indicate that the data-driven controller outperforms the PID baseline on this task. The PID controller exhibits a consistent lag behind the reference trajectory, a direct consequence of the uncompensated system delay. Furthermore, it struggles to overcome gravity, noticeably undershooting the target angle when lifting the arm (e.g., the left shoulder flexion in Figure~\ref{fig:data_driven_ctrl/compare_trajectory}(b)). In addition, the PID controller produces jerky motion due to stick-slip phenomena, which is especially visible during slower movements (e.g., from $10$–$14$~s in Figure~\ref{fig:data_driven_ctrl/compare_trajectory}(d)).

In contrast, the trained model tracks the reference trajectory closely. It successfully compensates for both the time delay and gravitational effects, and it effectively smooths out the stick-slip behavior, producing a smoother trajectory by having learned the system's nonlinear dynamics from the data. These visual findings are confirmed by RMSE values. For all four tested joints, the trained model achieves a lower tracking error than the PID controller. For the left shoulder flexion (joint 10), the data-driven controller reduced the RMSE from $8.79$ to $1.17$ in the normalized position unit defined in Section~\ref{sec:notation}, a reduction of nearly 87\%.

These experiments support the potential of a data-driven approach. By explicitly structuring the training data to account for system delay, even a simple MLP can learn an effective inverse model for this coupled, nonlinear pneumatic subsystem and outperform a standard PID controller. Section~\ref{sec:limitations} discusses what this evidence does and does not show.

\section{Discussion}
\label{sec:discussion}

The high reproducibility of the robot, confirmed in our experiments, suggests that it is a promising platform for human-robot interaction, although no interaction experiment with humans has yet been conducted. This section discusses the robot's potential for expressive motion and then turns to the controller study: why the arm was identified as a whole, the limitations of the study, and the directions they suggest.

\subsection{Implications for Expressive and Interactive Motion}
\label{sec:implications-for-motion}



The robot's capacity for high-speed motion, revealed in Section~\ref{sec:exp-max-velocity}, is a foundation for creating dynamic behaviors. This capacity could allow for quick, reflexive motions, such as flinching from a sudden, painful stimulus. In the context of human-robot interaction, such reflexes could make the robot seem more alive, potentially leading to more natural and empathetic interactions. The inherent compliance of the pneumatic system is also a major advantage for creating intense, oscillatory movements, like shaking the hands and shoulders to express excitement. While traditional robots with rigid, high-gear-ratio electric motors risk mechanical failure, the compressed air in the actuators naturally absorbs these rapid, opposing forces, protecting the mechanism from damage. This allows for a unique class of dynamic, emotional expressions that are difficult to achieve safely with other actuation methods.



Our experimental results from Section~\ref{sec:dynamic-properties} also provide practical guidelines for designing effective movements. The performance difference between the easiest (Pose E\textsubscript{P} or E\textsubscript{N}) and hardest (Pose H\textsubscript{P} or H\textsubscript{N}) conditions reveals a joint’s sensitivity to gravitational and inertial loads. For joints with high sensitivity, such as the shoulders, complex motions should be designed with consideration for the full arm's posture; for instance, lifting the shoulder is more efficient if the elbow is first flexed to reduce the arm's moment of inertia. The robot's high motion reproducibility, a key feature achieved by its robust design featuring pneumatic rotary actuators and cylinders with rigid bodies, was validated by the results in Section~\ref{sec:exp-reproducibility}. This property allows for identifying which joints are most consistent and therefore best suited for tasks requiring high precision. Conversely, joints with the highest maximum velocity are best for fast gestures. Finally, the minimum activation pressure results highlight the challenge of static friction, which must be considered when designing extremely slow and smooth movements to avoid jerky stick-slip motion.

\subsection{Interaction Effects Among Joints}
\label{sec:interaction-effects}

As described in Section~\ref{sec:data-collection}, the 4-DOF arm was identified simultaneously rather than joint by joint, and the interaction effects behind this choice should be clarified. The rigid-body coupling among the joints, that is, the inertial, Coriolis, centrifugal, and gravitational terms, can be computed from a dynamics model. However, the difficulty lies in the mapping from valve commands to joint torques, whose parameters change with the load imposed by the other joints. Experiments I-A and I-B in Sections~\ref{sec:exp-time-delay} and \ref{sec:exp-min-pressure} showed that both the transmission delay and the friction threshold depend on posture. In addition, the compressed air in the chambers should be treated as a configuration-dependent elastic element. A single-DOF inverse model identified in isolation therefore does not transfer to coordinated motion.

\subsection{Limitations of the Present Study}
\label{sec:limitations}

This study has several limitations. All measurements come from one physical robot, the dynamic properties were measured on the waist and left arm on the assumption that the two arms behave alike, and the controller was trained and tested on the left arm only. The difference between joints 6 and 12 in Section~\ref{sec:exp-reproducibility} shows that their trial-to-trial errors differ, but we did not pursue this variability further.

The controller is preliminary: a static three-layer MLP trained once on a fixed dataset, without a systematic hyperparameter search or an ablation of the lookahead period $\tau$, and with a constant delay assumed. It was compared only against a manually tuned PID controller and not against other learning-based or model-based controllers, so the results demonstrate feasibility rather than state-of-the-art performance. It also controls only the 4-DOF arm subsystem rather than the whole 13-DOF body.

The evaluation is also narrow. Tracking was tested on a single manually guided reference trajectory of moderate speed, with no payload change, external contact, or interaction with a human.

\subsection{Performance and Future Directions for the Data-Driven Controller}
\label{sec:controller-performance-future}


The successful trajectory tracking controller presented in Section~\ref{sec:data-driven-control} is a key contribution of this work. Its performance supports two premises. First, the robot system is sufficiently durable and reproducible for data-driven methods, as shown by its ability to collect 100 trials of continuous motion data without malfunction. Second, the proposed time-delay compensation strategy is effective. By training the MLP to look ahead by an interval $\tau$ derived from our experiments, the controller mitigated the inherent time delay and outperformed the traditional PID controller. This result supports further study of data-driven controllers for similar pneumatic robots.

Several directions for future work can be drawn from these results. The model was trained on moderately paced random motions, and its performance at the extremes of speed should be investigated. Tracking very fast movements is limited by the physical actuation delay, while tracking very slow movements is challenged by stick-slip effects. Furthermore, the controller's reliance on a fixed future period for the reference trajectory may not be optimal for dynamically generated or reflexive movements. Since the MLP is a static mapping, a future direction is to explore models that consider the history of states and inputs, such as a recurrent neural network, to capture the delay that varies with the actuator state and external loads. Extending the data-driven controller to the full 13-DOF robot is another important next step. A more advanced challenge is enabling the controller to adapt to changes in dynamics, such as when picking up an object, which would require online learning or adaptive control techniques to adjust the model in real time.

\section{Conclusion}
\label{sec:conclusion}
This paper presented the development and analysis of a compact, 13-DOF upper-body humanoid robot driven by pneumatic actuators. We conducted a series of experiments to systematically characterize its dynamic properties, which revealed a system that is complex and highly nonlinear, yet robust and reproducible within the tested conditions. A key challenge identified and quantified through these experiments was a significant time delay, primarily due to the long pneumatic transmission lines.

To address these control challenges, we proposed and implemented a preliminary data-driven controller. The core of our approach was to train a multilayer perceptron as an inverse dynamics model, using a data structure specifically designed to compensate for the known system delay by incorporating future desired states into the model's input. When evaluated on an unseen, manually guided trajectory, this controller achieved lower tracking errors than a standard PID controller, mitigating both the time delay and configuration-dependent gravitational effects.

This work shows that a data-driven approach is a viable strategy for controlling a coupled 4-DOF pneumatic arm of this robot and a promising direction for the full 13-DOF system. Within the tested conditions, it also establishes this robot as a durable and reproducible platform for further work on adaptive control and on expressive motion for human-robot interaction. The limitations of the study, in particular that only one robot was tested and that the controller was compared only with a PID baseline, are discussed in Section~\ref{sec:limitations}.

\section*{Funding}

This work was supported in part by the project, JPNP16007, commissioned by the New Energy and Industrial Technology Development Organization (NEDO), and in part by JSPS KAKENHI Grant-in-Aid for Transformative Research Areas Grant Number JP 25H01236.

\bibliographystyle{tfnlm}
\bibliography{affetto}

\end{document}